\documentclass[conference]{IEEEtran}
\IEEEoverridecommandlockouts
\usepackage{cite}
\usepackage{amsmath,amssymb,amsfonts}
\usepackage{algorithmic}
\usepackage{graphicx}
\usepackage{xspace}
\usepackage{textcomp}
\usepackage{xcolor}
\usepackage{adjustbox}  %mainly for table shrink
\usepackage{algorithm}
\usepackage{algorithmic}
\usepackage{multirow}
\usepackage{booktabs}
\usepackage{color}

\usepackage[a4paper, total={185mm,246mm}]{geometry}
\def\BibTeX{{\rm B\kern-.05em{\sc i\kern-.025em b}\kern-.08em
    T\kern-.1667em\lower.7ex\hbox{E}\kern-.125emX}}
\begin{document}

\addtolength{\textheight}{0.25in}
\addtolength{\textwidth}{0.01in}
% Redefine the \subsection command to include \vspace{-2pt}
\let\oldsubsection\subsection

\newcommand{\ours}{OPTIMIS\xspace}

\newcommand{\Wei}[1]{{\color{blue} #1}}	%color text for comments
\newcommand{\toread}[1]{{\color{magenta} Read paper to rewrite: #1}}	%color text for comments
\newcommand{\help}[1]{{\color{red} YF: #1}}	%color text for comments

% Math commands by Wei W. Xing (wxing.me) by merging Math commands by Thomas Minka and AAAI's math_commands.tex

% ref
\def\Eqref#1{Eq.~\eqref{#1}}
\def\Algref#1{Algorithm~\ref{#1}}
\def\Figref#1{Fig.~\ref{#1}}

\newcommand{\var}{{\rm Var}}
\newcommand{\Tr}{^{\rm Tr}}
\newcommand{\vtrans}[2]{{#1}^{(#2)}}
\newcommand{\kron}{\otimes}
\newcommand{\schur}[2]{({#1} | {#2})}
\newcommand{\schurdet}[2]{\left| ({#1} | {#2}) \right|}
\newcommand{\had}{\circ}
\newcommand{\diag}{{\rm diag}}
\newcommand{\invdiag}{\diag^{-1}}
\newcommand{\rank}{{\rm rank}}
% careful: ``null'' is already a latex command
\newcommand{\nullsp}{{\rm null}}
\newcommand{\tr}{{\rm tr}}
\renewcommand{\vec}{{\rm vec}}
\newcommand{\vech}{{\rm vech}}
\renewcommand{\det}[1]{\left| #1 \right|}
\newcommand{\pdet}[1]{\left| #1 \right|_{+}}
\newcommand{\pinv}[1]{#1^{+}}
\newcommand{\erf}{{\rm erf}}
\newcommand{\hypergeom}[2]{{}_{#1}F_{#2}}

% boldface characters
\renewcommand{\a}{{\bf a}}
\renewcommand{\b}{{\bf b}}
\renewcommand{\c}{{\bf c}}
\renewcommand{\d}{{\rm d}}  % for derivatives
\newcommand{\e}{{\bf e}}
\newcommand{\f}{{\bf f}}
\newcommand{\g}{{\bf g}}
\newcommand{\h}{{\bf h}}
%\newcommand{\k}{{\bf k}}
% in Latex2e this must be renewcommand
\renewcommand{\k}{{\bf k}}
\newcommand{\m}{{\bf m}}
\newcommand{\n}{{\bf n}}
\renewcommand{\o}{{\bf o}}
\newcommand{\p}{{\bf p}}
\newcommand{\q}{{\bf q}}
\renewcommand{\r}{{\bf r}}
\newcommand{\s}{{\bf s}}
\renewcommand{\t}{{\bf t}}
\renewcommand{\u}{{\bf u}}
\renewcommand{\v}{{\bf v}}
\newcommand{\w}{{\bf w}}
\newcommand{\x}{{\bf x}}
\newcommand{\y}{{\bf y}}
\newcommand{\z}{{\bf z}}
%s\newcommand{\l}{\boldsymbol{l}}
\newcommand{\A}{{\bf A}}
\newcommand{\B}{{\bf B}}
\newcommand{\D}{{\bf D}}
\newcommand{\E}{{\bf E}}
\newcommand{\F}{{\bf F}}
\renewcommand{\H}{{\bf H}}
\newcommand{\I}{{\bf I}}
\newcommand{\J}{{\bf J}}
\newcommand{\K}{{\bf K}}
\renewcommand{\L}{{\bf L}}
\newcommand{\M}{{\bf M}}
\newcommand{\N}{\mathcal{N}}  % for normal density
\newcommand{\MN}{\mathcal{MN}} 
\newcommand{\Acal}{\mathcal{A}}
\newcommand{\Ocal}{\mathcal{O}}
\newcommand{\Dcal}{\mathcal{D}}
\newcommand{\Ycal}{\mathcal{Y}}
\newcommand{\Zcal}{\mathcal{Z}}
\newcommand{\Fcal}{\mathcal{F}}
\newcommand{\Vcal}{\mathcal{V}}
\newcommand{\Lcal}{\mathcal{L}}
\newcommand{\Tcal}{\mathcal{T}}
\newcommand{\Gcal}{\mathcal{G}}
\newcommand{\Hcal}{\mathcal{H}}
\newcommand{\Scal}{\mathcal{S}}
\newcommand{\Xcal}{\mathcal{X}}

\renewcommand{\O}{{\bf O}}
\renewcommand{\P}{{\bf P}}
\newcommand{\Q}{{\bf Q}}
\newcommand{\R}{{\bf R}}
\renewcommand{\S}{{\bf S}}
\newcommand{\T}{{\bf T}}
\newcommand{\V}{{\bf V}}
\newcommand{\W}{{\bf W}}
\newcommand{\X}{{\bf X}}
\newcommand{\Y}{{\bf Y}}
\newcommand{\Z}{{\bf Z}}
\newcommand{\Mcal}{{\mathcal{M}}}
\newcommand{\Wcal}{{\mathcal{W}}}
\newcommand{\Ucal}{{\mathcal{U}}}

% this is for latex 2.09
% unfortunately, the result is slanted - use Latex2e instead
%\newcommand{\bfLambda}{\mbox{\boldmath$\Lambda$}}
% this is for Latex2e
\newcommand{\bfLambda}{\boldsymbol{\Lambda}}

% Yuan Qi's boldsymbol
\newcommand{\bsigma}{\boldsymbol{\sigma}}
\newcommand{\balpha}{\boldsymbol{\alpha}}
\newcommand{\bpsi}{\boldsymbol{\psi}}
\newcommand{\bphi}{\boldsymbol{\phi}}
\newcommand{\boldeta}{\boldsymbol{\eta}}
\newcommand{\Beta}{\boldsymbol{\eta}}
\newcommand{\btau}{\boldsymbol{\tau}}
\newcommand{\bvarphi}{\boldsymbol{\varphi}}
\newcommand{\bzeta}{\boldsymbol{\zeta}}
\newcommand{\bepsilon}{\boldsymbol{\epsilon}}

\newcommand{\blambda}{\boldsymbol{\lambda}}
\newcommand{\bLambda}{\mathbf{\Lambda}}
\newcommand{\bOmega}{\mathbf{\Omega}}
\newcommand{\bomega}{\mathbf{\omega}}
\newcommand{\bPi}{\mathbf{\Pi}}

\newcommand{\btheta}{\boldsymbol{\theta}}
\newcommand{\bpi}{\boldsymbol{\pi}}
\newcommand{\bxi}{\boldsymbol{\xi}}
\newcommand{\bSigma}{\boldsymbol{\Sigma}}

\newcommand{\bgamma}{\boldsymbol{\gamma}}
\newcommand{\bGamma}{\mathbf{\Gamma}}

\newcommand{\bmu}{\boldsymbol{\mu}}
\newcommand{\1}{{\bf 1}}
\newcommand{\0}{{\bf 0}}

\newcommand{\bs}{\backslash}

 \newcommand{\notS}{{\backslash S}}
 \newcommand{\nots}{{\backslash s}}
 \newcommand{\noti}{{\backslash i}}
 \newcommand{\notj}{{\backslash j}}
 \newcommand{\nott}{\backslash t}
 \newcommand{\notone}{{\backslash 1}}
 \newcommand{\nottp}{\backslash t+1}

\newcommand{\notk}{{^{\backslash k}}}
\newcommand{\notij}{{^{\backslash i,j}}}
\newcommand{\notg}{{^{\backslash g}}}
\newcommand{\wnoti}{{_{\w}^{\backslash i}}}
\newcommand{\wnotg}{{_{\w}^{\backslash g}}}
\newcommand{\vnotij}{{_{\v}^{\backslash i,j}}}
\newcommand{\vnotg}{{_{\v}^{\backslash g}}}
\newcommand{\half}{\frac{1}{2}}
\newcommand{\msgb}{m_{t \leftarrow t+1}}
\newcommand{\msgf}{m_{t \rightarrow t+1}}
\newcommand{\msgfp}{m_{t-1 \rightarrow t}}

\newcommand{\proj}[1]{{\rm proj}\negmedspace\left[#1\right]}
\newcommand{\argmin}{\operatornamewithlimits{argmin}}
\newcommand{\argmax}{\operatornamewithlimits{argmax}}

\newcommand{\dif}{\dfrac{\dfrac{\mathrm}{den}}{den}{d}}
\newcommand{\abs}[1]{\lvert#1\rvert}
\newcommand{\norm}[1]{\lVert#1\rVert}

%miscellaneous symbols
%\newcommand{\ie}{{{\em i.e.,}}\xspace}
%\newcommand{\ie}{{\textit{i.e.,}}\xspace}
%\newcommand{\eg}{{\textit{e.g.,}}\xspace}
%\newcommand{\etc}{{\textit{etc.}}\xspace}
\newcommand{\ie}{{i.e.,}\xspace}
\newcommand{\eg}{{e.g.,}\xspace}
\newcommand{\etc}{{etc.}\xspace}

\newcommand{\EE}{\mathbb{E}}
\newcommand{\dr}[1]{\nabla #1}
\newcommand{\VV}{\mathbb{V}}
\newcommand{\sbr}[1]{\left[#1\right]}
\newcommand{\rbr}[1]{\left(#1\right)}
\newcommand{\cmt}[1]{}

\newcommand{\bi}{{\bf i}}
\newcommand{\bj}{{\bf j}}
\newcommand{\bK}{{\bf K}}
\newcommand{\Vtr}{\mathrm{Vec}}

\newcommand{\cov}{{\rm Cov}}	%if {\rm Cov} \rm effects all components

\newtheorem{Proposition}{Proposition}
\newtheorem{Lemma}{Lemma}
\newtheorem{Corollary}{Corollary}
\newtheorem{Remark}{Remark}
\newtheorem{Assumption}{Assumption}
\newtheorem{Property}{Property}

\newcommand{\RR}{\mathbb{R}}
\newcommand{\KL}{\mathrm{KL}}

%new command for physical journal by wxing
%use 
%\newcommand{\bpsi}{\boldsymbol{\psi}}
\newcommand{\bPsi}{\boldsymbol{\Psi}}
\newcommand{\bXi}{\boldsymbol{\Xi}}
\newcommand{\btx}{\textbf{\textit{x}}}
\newcommand{\bty}{\textbf{\textit{y}}}
\newcommand{\btz}{\textbf{\textit{z}}}
\newcommand{\btk}{\textbf{\textit{k}}}

\newcommand{\bupsilon}{\boldsymbol{\upsilon}}

\newcommand{\GP}{\mathcal{GP}}
\newcommand{\TGP}{\mathcal{TGP}}
\newcommand{\TNcal}{\mathcal{TN}}

% \title{High Dimensional Yield Estimation Using an Adaptive Normalizing Flows Sampling Method*\\
\title{Approaching the Key Barrier For High-Dimensional Yield Estimation Using Manifold Learning}
\title{OPTIMIS: Optimal Manifold Important Sampling For High-Dimensional Yield Estimation}
% in homage to Breaking the Simulation Barrier: SRAM Evaluation Through Norm Minimization
\title{Constructing the Simulation Barrier: SRAM High-Dimensional Yield Evaluation Through Optimal Manifold}
\title{Optimal Important Sampling For High-Dimensional SRAM: From Points to Manifold}
\title{Constructing the Yield Barrier: High-Dimensional SRAM Evaluation Through Optimal Manifold}
\title{Seeking the Yield Barrier: High-Dimensional SRAM Evaluation Through Optimal Manifold \vspace*{-0.3cm}
\thanks{This work is supported by Fundamental Research Funds for the Central Universities; experiments are supported by Primarius Technologies Co.,Ltd.}
}
% }
% \author{
% 	\IEEEauthorblockN{Yanfang Liu\IEEEauthorrefmark{1}, Guohao Dai\IEEEauthorrefmark{2},  and Wei W. Xing\IEEEauthorrefmark{1}} 
% 	\IEEEauthorblockA{\IEEEauthorrefmark{1} School of Integrated Circuit Science and Engineering, Beihang University, Beijing, China 100191}
%  	\IEEEauthorblockA{\IEEEauthorrefmark{2} College of Mechatronics and Control Engineering, Shenzhen University, Shenzhen, China}
% 		% {wxing@buaa.edu.cn}
%   \thanks{Yanfang Liu and Guohao Dai contribute equally to this work}
%   \thanks{Wei W. Xing (wxing@buaa.edu.cn) is the corresponding author}
%   }
\author{\IEEEauthorblockN{Yanfang Liu\IEEEauthorrefmark{2} \thanks{\IEEEauthorrefmark{2}Both authors contributed equally to this research.}}
\IEEEauthorblockA{\textit{School of Integrated Circuit Science } \\
\textit{and Engineering, Beihang University}\\
Beijing, China \\
liuyanfang@buaa.edu.cn}
\and
\IEEEauthorblockN{Guohao Dai\IEEEauthorrefmark{2}}
\IEEEauthorblockA{\textit{College of Mechatronics and Control
} \\
\textit{Engineering, Shenzhen University}\\
Shenzhen, China \\
daiguohao2019@email.szu.edu.cn}
\and
\IEEEauthorblockN{Wei W. Xing\IEEEauthorrefmark{1}
\thanks{\IEEEauthorrefmark{1}Corresponding author.}
% \thanks{\IEEEauthorrefmark{3}Also affiliated with Beihang Hangzhou Innovation Institute Yuhang, Hangzhou, China.}
}
\IEEEauthorblockA{\textit{School of Integrated Circuit Science } \\
\textit{and Engineering, Beihang University}\\
Beijing, China \\
wxing@buaa.edu.cn}
% \vspace*{-1.2cm}
}

\maketitle
% \vspace*{-1cm}
% \vspace*{-1.2cm}

% \IEEEaftertitletext{\vspace{-2cm}} % add 20pt of vertical space after the title

\begin{abstract}
Being able to efficiently obtain an accurate estimate of the failure probability of SRAM components has become a central issue as model circuits shrink their scale to submicrometer with advanced technology nodes. 
% The practical includes the multiple failure regions issues and ``curse of dimensionality''.
% 
In this work, we revisit the classic norm minimization method. We then generalize it with infinite components and derive the novel optimal manifold concept, which bridges the surrogate-based and importance sampling (IS) yield estimation methods.
We then derive a sub-optimal manifold, optimal hypersphere, which leads to an efficient sampling method being aware of the failure boundary called onion sampling.
Finally, we use a neural coupling flow (which learns from samples like a surrogate model) as the IS proposal distribution.
% and propose an approximated solution, named onion sampling, to efficiently draw samples from the ground-truth failure probability.
% Based on these samples, we use a neural coupling flow a to sequentially becomes the optimal proposal distribution
% Finally, we use a normalization flow (as in surrogate-based methods) that sequentially approximates the ground truth failure distribution in an IS method to act 
% To harness the knowledge carried by samples about the optimal manifold, we 
% use a normalization flow  and 
These combinations give rise to a novel yield estimation method, named Optimal Manifold Important Sampling (OPTIMIS), which keeps the advantages of the surrogate and IS methods to deliver state-of-the-art performance with robustness and consistency, with up to 3.5x in efficiency and 3x in accuracy over the best of SOTA methods in High-dimensional SRAM evaluation.

\end{abstract}

\begin{IEEEkeywords}
Yield Analysis, Importance Sampling, Normalization Flow
\end{IEEEkeywords}

% \Wei{
% \section*{Flow}
% \begin{itemize}
%     \item Intro
%     \item Problems with surrogate-based methods
%     \item Problems with IS-based method; SOTA: vMF; Center problem: OSV
%     \item Our Solution, manifold ridge uisng NF. Motivation+derivation+anti mode collapse +(some comparison to vMF and mixture GP) + illustration figures
%     % \item Resolve the challenge of N samples: novel pre-sampling
%     \item *complexity+unbias estimation
%     \item connection to surrogate-based methods
% \end{itemize}
% Some future works: soften I(x); conditional NF
% highlight of this work:
% \begin{itemize}
%   \item extension of the norm minimization, from points to hyperplanes 
%   \item utilization of model ML: NF to efficiently draw samples from a complex manifold based distributions.
%   \item Connection to surrogate model
%   \item Super efficiency
% \end{itemize}
% }

\section{Introduction}
As the technology of integrated circuits develops, microelectronic devices shrink their scale to submicrometer, which makes random process variations, \eg intra-die mismatches, doping fluctuation, and threshold voltage variation, crucial factors to be considered in a circuit design.
The situation gets worse in modern circuit designs, where some cells can be replicated millions of times in a circuit, \eg in an SRAM cell array.
A cornerstone to resolving the increasing concern of yield is the development of efficient yield estimation methods, which provide an accurate and fast failure probability estimation for a given circuit design under specific process variations.

Monte Carlo (MC) is the golden standard baseline, and it is commonly utilized to estimate the yield across industry and academia. 
In a nutshell, MC runs SPICEs (Simulation Program with Integrated Circuit Emphasis) with parameters sampled from the process variation distribution millions of times and counts the number of failures to deliver accurate estimation.
Obviously, MC is computationally expensive and easily becomes infeasible for problems for low-yield problems, which is rather common in modern circuit designs, \eg the yield of a 65nm SRAM cell array can be $10^{-5}$.
% It will
% However, for high-dimensional circuits, the MC method is extremely inefficient, because high-dimensional SPICE simulation is expensive and time-consuming. 
% However, due to the robust modern design of circuits, the failure probability of circuits is extremely low (usually smaller than $10^{-6}$), which is difficult to estimate.

% yield estimation methods, which can accurately predict the failure probability of a circuit design.
% The reliability of circuits has been a growing concern because the failure of the circuits can be induced by random process variation during manufacturing,  Given process variations, it's crucial to estimate the yield of circuits. 

% \Wei{add convergence rate of MC}
To improve the efficiency of yield estimation, importance-sampling (IS)-based methods have been proposed. 
Instead of drawing samples from the default normal distribution, IS methods draw samples from a proposal distribution, which should be designed to approximate the oracle failure distribution.
Thus, most efforts tried to design a proposal distribution that can approximate the failure distribution well.
For instance, \cite{MNIS} shifts the sampling centroid of the normal distribution to the closest failure point as the proposal distribution, which is {well-known} as norm minimization (NM).
Based on the shifting idea, \cite{HSCS} proposes to sample from multiple failure regions using a hyperspherical clustering method. 
% according to a constructed distribution whose mean vector is shifted to the failure regions, thus accelerating the discovery of failure samples. \cite{MNIS} shifts the sampling centroid to a single failure region. 
Instead of relying on a static proposal distribution, \cite{AIS} proposes an adaptive importance sampling (AIS) to update the shifted distribution as more samples are collected.
To better fit the failure distribution in a high-dimensional space, \cite{ACS} samples from multiple regions clustered by multi-cone clustering and sequentially updates its proposal distribution.
AIS is further enhanced by \cite{NGAIS} by introducing a mixture of von Mises-Fisher distributions to replace the standard normal distribution.
The IS methods are robust and simple to implement, making them popular in the industry. Nonetheless, they still require a large number of SPICE simulations and can not incorporate coming knowledge of new samples to update their models (\eg the proposal distribution and/or its family).
% MORE 
% Their convergence is irrelevant to the dimensionlity.
% However, the importance-sampling-based methods mentioned above require a large number of presampling to seek enough failure samples to train the distribution. And these methods are inefficient to estimate the yield in high-dimensional circuits.

Another main path to efficient yield estimation is utilizing powerful machine learning (ML) to build a data-driven surrogate model to approximate the unknown indicator function and use active learning to sequentially reduce prediction error.
% The selection of the surrogate model is crucial. 
\cite{yin2022efficient} uses a Gaussian process (GP) to approximate the underlying performance function and an entropy reduction method for active learning.
Based on the same updating scheme, \cite{ASDK} replaces the GP with a nonlinear-correlated deep kernel method with feature selection to identify the determinant features to focus on.
Instead of using a two-stage approach that potentially introduces bias, \cite{LRTA} uses a low-rank tensor approximation to the polynomial chaos expansion (PCE) to approximate the performance function.
Deep learning (\eg RBF network) can {(also be)} utilized in combination with importance sampling to compute the yield \cite{AMSV}.
Despite their success, the surrogate methods inevitably suffer from the ``curse of dimensionality''. More specifically, the high dimensionality (which is quite common in SRAM circuits) makes it challenging to compute the integration over the domain and data demanding to train the surrogate model, which can defeat the purpose of introducing a surrogate model. Another critical problem with surrogate-based methods is the highly nonlinear optimization problem involved in the surrogate model training, which, 
if not done right, can lead to a wrong surrogate model and thus a wrong yield estimation, a disaster the industry can not afford.

% because (1) computation of yield requires expensive integration and (2)
% they still require expensive integration over the domain to calculate yield and (2) the 
% which is a major challenge in high-dimensional circuits.
% \cite{AMSV} combines surrogate with , using an RBF network to replace the simulator.

% Another way to improve the efficiency of yield estimation is resorting to surrogate model. Instead of using a expensive SPICE simulator to evaluate samples, surrogate-modelling-based methods train a data-driven model to replace the simulator, thus avoiding the frequent calls to the simulator. \cite{LRTA} constructs a surrogate model based on low-rank tensor approximation. combines surrogate with importance sampling, using an RBF network to replace the simulator. However, the existing surrogate methods suffer from the ``curse of dimensionality''. The estimation result is highly dependent to the accuracy of surrogate. To get an accurate surrogate, the number of required training data explodes as the dimensions of the circuit increase, which is utterly infeasible.

In this work, we aim to combine the advantages of IS and surrogate methods to deliver an efficient and, most importantly, robust yield estimation.
To this end, we first revisit the classic NM method (based on which many methods are proposed); we generalize it with infinite components and derive the optimal manifold, which reveals the close connection between IS and surrogate methods and serves as a guideline in designing the proposal distribution in IS methods.
% 
% serves as a guideline in designing the proposal distribution in IS methods. Another interesting discover is that the optimal manifold also reveals the closed connection between IS and surrogate methods.
% 
Based on the optimal manifold, we propose a sub-optimal solution, optimal hypersphere, and derive onion sampling, which provides efficient and robust samples from the failure distribution.
Finally, we introduce neural spline flows (NSF) as the proposal distribution; it sequentially approximates the truth failure probability with more samples collected (as in a surrogate method) to deliver efficient sampling.
% To enable model update and active learning in surrogate methods, we use a normalization flow (NF) to sequentially approximate the truth failure distribution with more samples are collected.
This combination gives rise to a novel sampling method, named \textbf{Opti}mal \textbf{M}anifold \textbf{I}mportant \textbf{S}ampling (OPTIMIS),
% Optimal Manifold Important Sampling (OPTIMIS), 
which absorbs the advantages of surrogate-based and IS-based methods to deliver state-of-the-art (SOTA) performance with robustness and efficiency.
%  with up to 2.5x speedup over the SOTA methods in High-dimensional SRAM evaluation.
% 
% any complex distribution
% To resolve these issues, we seek a manifold that approximates $q^*(\x)$ using , which is expect to approximate any complex distribution using the power of deep learning and massive parallel computing through GPU. Most Importantly, this method can automatically update itself as more samples are collected to reveal the true failure boundary (as is done in a surrogate-based method).
The novelty of this work includes:
\begin{enumerate}
  \item Optimal manifold: a generalization of the classic NM.
  % \item Bridging the gap between IS and surrogate methods.
  \item Onion sampling, which can efficiently sample from the failure distribution
  \item \ours, combining onion sampling and NSF to deliver SOTA  yield estimation; which is as efficient as the surrogate methods and as robust as the IS methods.
  \item The superiority is valid with SRAM circuits with up to 1093 dimensions, ablation study, and robustness test.
  % of for more than 1000-dimensional SRAM circuit
  % to update the proposal distribution as in surrogate methods.
\end{enumerate}

% To handle the high-dimensional yield analysis challenge, we propose an adaptive normalization flow sampling method. Our major contributions are: 
% (1) A novel presampling method to identify failure regions. The method explores the whole variation parameter space and discovers the most promising failure regions. 
% (2) We use a novel as normalization flow(NF) model as IS distribution. The NF model is able to fit the failure region in high-dimensional variation parameter space. 
% (3) We utilize the adaptive scheme in \cite{NGAIS} to calibrate the IS distribution iteratively. 
% (4) The experimental results shows that \ours is faster than other SOTA models without losing accuracy in high-dimensional circuit cases. 
% The rest of this paper is arranged 

\section{Background}
\subsection{Problem Definition}
Denote $\x=[x^{(1)},x^{(2)},\cdots,x^{(D)}]^T \in \Xcal$ as the variation process parameter, and $\Xcal$ the variation parameter space. 
$\Xcal$ is generally a high-dimensional space (\ie large $D$);
%  whereas each variable in $\x$ is mutually independent.
Each variable in $\x$ denotes the variation parameters of a circuit during manufacturing, \eg length or width of PMOS and PNOS transistors. 
In general, $\x$ are considered mutually independent Gaussian distributed, 
% which follows the probability density function (PDF):
% $ p(\x) = (2\pi)^{\frac{D}{2}} \prod_{d=1}^D \exp \left(- (x^{(d)})^2 /2 \right).$
$ p(\x) = (2\pi)^{\frac{D}{2}}\exp \left(- ||\x||^2 /2 \right).$
% \begin{equation}
%   p(\x) = (2\pi)^{\frac{D}{2}} \prod_{d=1}^D \exp \left(- (x^{(d)})^2 /2 \right).
% \end{equation}
% In real-life manufacturing process of circuits, each variable in $\x$ is mutually independent, and $\x$ follows a multivariate Gaussian distribution. So after normalization of $\x$, we have the probability density function (PDF) 
% \[
%   p(\x) = \prod_i^d \exp \left(- (x^{(i)})^2 /2 \right)/{\sqrt{2 \pi}}.
%   \]
% $ p(\x) = \prod_i^d \exp \left(- (x^{(i)})^2 /2 \right)/{\sqrt{2 \pi}}$.
Given a specific value of $\x$, we can evaluate the circuit performance $\y$ (\eg  memory read/write time and amplifier gain) through SPICE simulation, $ \y=\f(\x)$,
% \begin{equation}
%     \y=\f(\x)
% \end{equation}
where $\f(\cdot)$ is the SPICE simulator, which is considered an expensive and time-consuming black-box function;
$\y=[y^{(1)},y^{(2)},\cdots,y^{(K)}]^T$ are the collections of circuit performance based on the simulations.
% where $\boldsymbol{f}$ is considered as a black-box function of SPICE simulator. 
When $K$ metrics are all smaller than or equal to their respective thresholds (predefined by designers) $\boldsymbol{t}$, \ie $y^{(k)} \leq t^{(k)}$ for $k=1,\cdots,K$, the circuit is considered as a successful design; otherwise, it is a failure one.
% where $\boldsymbol{t}=[t^{(1)},t^{(2)},\cdots,t^{(K)}]^T$ are the thresholds of circuit performance predefined by designers, we can classify the corresponding circuit for $\x$ a successful  design. Otherwise, the circuit is considered a failure design.
We use failure indicator $I(\x)$, which is 1 if $\x$ lead to a failure design and 0 otherwise, to denote the failure status of a circuit.
% 
% a circuit is considered as failed if $\y^{(k)} \geq \boldsymbol{t}^{(1)}$ for any $k \in \{1,\cdots,K\}$.
% When values of $K$ metrics in $\y$ are all smaller than or equal to their respective thresholds $\boldsymbol{t}$, \ie $y^{(k)} \leq t^{(k)}$ for $k=1,\cdots,K$, we can classify the corresponding $\x$ of the $\y$ as a successful sample. Otherwise, the  corresponding $\x$ is classified as a failure sample. We can further introduce the failure indicator function:
% \begin{equation}
%         I(\x)=\left\{
%           \begin{aligned}
%           & 1, \; y^{(k)} > t^{(k)} \;  \mathrm{for} \; \exists k\\ 
%           & 0, \; y^{(k)} \leq t^{(k)}\; \mathrm{if} \; \forall k\\
%           \end{aligned}
%           \right.
% \end{equation}
Finally, the ground-truth failure rate $\hat{P}_f $ is:
% yield is the probability of failure samples:
% we can estimate the failure probability $P_f$ by:
$\hat{P}_f = \int_\mathcal{X} I\left( \x \right) p(\x) d\x.$
% \begin{equation}
% \label{Pfintegral}
%     \hat{P}_f = \int_\mathcal{X} I\left( \x \right) p(\x) d\x% = \EE_{p}[I(\x)]
% \end{equation}
    
\subsection{Monte Carlo and Importance Sampling}
The direct calculation of the yield is intractable due to the unknown $I(\x)$.
% the high dimensionality of $\Xcal$ and the expensive SPICE simulator. 
A common approach to estimate the failure probability is Monte Carlo (MC), which is easily implemented by sampling $\x_i$ from $p(\x)$ and evaluating the failure probability by the ratio of failure samples to total samples. More specifically, $\hat{P}_f $ is approximated by
% is intractable because $\f(\cdot)$ is a black-box function and the analytic solution of \Eqref{Pfintegral} is unachievable. Therefore, we resort to brute-force Monte Carlo method. The MC method draws $N$ samples according to PDF $p(\x)$. 
% The estimated failure rate $P_f$ is 
$P_f =\frac{1}{N} \sum_{i=1}^N I(\x_i),$
% \begin{equation}
%   P_f =\frac{1}{N} \sum_{i=1}^N I(\x_i),
% \end{equation}
where $\x_i$ is the $i$-th sample from $p(\x)$, and $N$ is the number of samples.
When $N \rightarrow \infty$, $P_f \rightarrow \hat{P}_f $.
% The estimated result of the MC method is regarded as the ground truth of the circuit failure probability. 
% 
% Despite the simplicity of MC, 
To obtain an estimate of $1- \varepsilon$ accuracy with $1-\delta$ confidence, $N \approx \frac{ \log(1/\delta)}{\varepsilon^2 \hat{P}_f }$ is required.
For a modest $90\%$ accuracy $(\varepsilon=0.1)$ with $90\%$ confidence $(\delta=0.1)$, we need $N \approx 100/\hat{P}_f$ samples, which is infeasible in practice for small $\hat{P}_f$, says, $10^{-5}$.
We can also see this intuitively from the fact that it requires averagely $1/\hat{P_f}$ samples just to observe a failure event.

% However, MC is very low-efficient.
% For yield estimation where the failure probability is very small, the number of samples required to achieve a certain accuracy is very large.

% To achieve a accurate estimation, The brute-force MC method require a large number of samples and SPICE simulation runs, which is extremely expensive in high-dimensional circuits. 
% MC takes too many samples to estimate small p because it requires 1/p samples on average just to observe the event A (or failure) even once !

Instead of drawing samples from $p(\x)$, the IS methods draw samples from a proposal distribution $q(\x)$ and estimate
% \begin{equation}
% \begin{split}
%     P_f 
%     &= \int_\mathcal{X} I(\boldsymbol{f}(\x)) p(\x) d\x
%     = \int_\mathcal{X} \frac{I(\boldsymbol{f}(\x)) p(\x) q(\x)} { q(\x) } d\x \\
%     &= \EE_{q(\x)}[\frac{I(\boldsymbol{f}(\x))p(\x)}{q(\x)}]
%     \approx \frac{1}{N} \sum_{i=1}^N \frac{I( \boldsymbol{f} (\x_i)) p(\x_i)} {g(\x_i)}.
% \end{split}
% \end{equation}
\begin{equation}
  \label{eq:IS}
  \begin{aligned}
    P_f  &= \int_\mathcal{X} \frac{I(\x) p(\x)} { q(\x) }  q(\x) d\x 
    \approx \frac{1}{N} \sum_{i=1}^N \frac{I(\x_i) p(\x_i)} {q(\x_i)},
    % = \sum_{i=1}^N {I(\x_i) w(\x_i)},
  \end{aligned}
\end{equation}
where $\x_i$ are samples drawn from $q(\x)$ and are used to approximate the integral as in MC. For convenience, we define the importance weight $w(\x)=\frac{p(\x)}{q(\x)}$.
\Eqref{eq:IS} is more efficient than MC if $q(\x)$ is chosen carefully.
% to approximate $I(\x)p(\x)/\hat{P}_f$, which is certainly problem dependent.

% In this work, we put forth a novel way to efficiently approximate the ground-truth distribution in an active learning framework.

% Addressing this challenge requires problem dependent creative solution. 
% In this paper, we shall put-forth a novel method, based on a new norm minimization principle, to address this challenge for a class of problems. We establish the effectiveness of our solution in the context of evaluating the probability of failure of SRAM elements in Section IV. Before presenting the algorithm in Section III, we establish some useful definitions.
% 
% In such a scenario, we can indeed speed-up the MC method by first finding pˆ quickly using samples of X and then using the knowledge of function w appropriately. This is the key idea behind Importance Sampling (IS).
% 
% where  With $N$ samples drawn from $q(\x)$, we can compute $P_f$ by:
% $q(\x)$ is a distribution distorted to failure regions in variation parameter space. 
% 
% The estimation from IS is unbiased and asymptotic w.r.t $\hat{P_f}$.

\subsection{Norm Minimization}
% \Wei{Can be removed}
One of the foundation work in IS for yield is the Norm Minimization (NM \cite{MNIS}, aka \textit{optimal shift vector}), which samples from a normal distribution centered at $\bmu^*$, 
% which solves the following optimization problem:
\begin{equation}
  \label{eq:OSV}
  \bmu^* = \argmin ||\x||^2  \quad \mathrm{s.t.} \quad I(\x) = 1,
\end{equation}
where $||\x||^2 = \sum_{d=1}^D (x^{(d)})^2$ is the Euclidean norm.
% \begin{equation}
%   \label{eq:OSV}
%   \bmu^* = \min_{\x} \sum_{d=1}^D |\frac{(x^{(d)})}{s^{(d)}}|^2  \quad \mathrm{for} \quad I(\x) = 1,
% \end{equation}
% where $s^{(d)}$ is the variance at $d$-th dimension for the proposed Gaussian distribution.
% The IS methods require a constructed PDF $q(\x)$ to draw samples. The optimal shift vector $\boldsymbol{\bmu}$ is a key parameter in the IS methods. The optimal shift vector $\boldsymbol{\bmu}$ is defined as:
% \begin{equation}
%     \boldsymbol{\bmu} = \arg \min_{\boldsymbol{\bmu}} \left\{ \frac{1}{N} \sum_{i=1}^N \frac{I( \boldsymbol{f} (\x_i)) p(\x)} {q(\x)} \right\}
% \end{equation}
% where $q(\x) = \prod_i^d \exp \left(- (x^{(i)}-\bmu^{(i)})^2 /2 \right)/{\sqrt{2 \pi}}$. The optimal shift vector $\boldsymbol{\bmu}$ is the solution of the following optimization problem:

% \Wei{problem with OSV}

\begin{figure*}[!ht]
  \centering
  \begin{minipage}[t]{0.9\linewidth}
  \centering
  \vspace*{-6pt}       
  \includegraphics[width=0.19\linewidth]{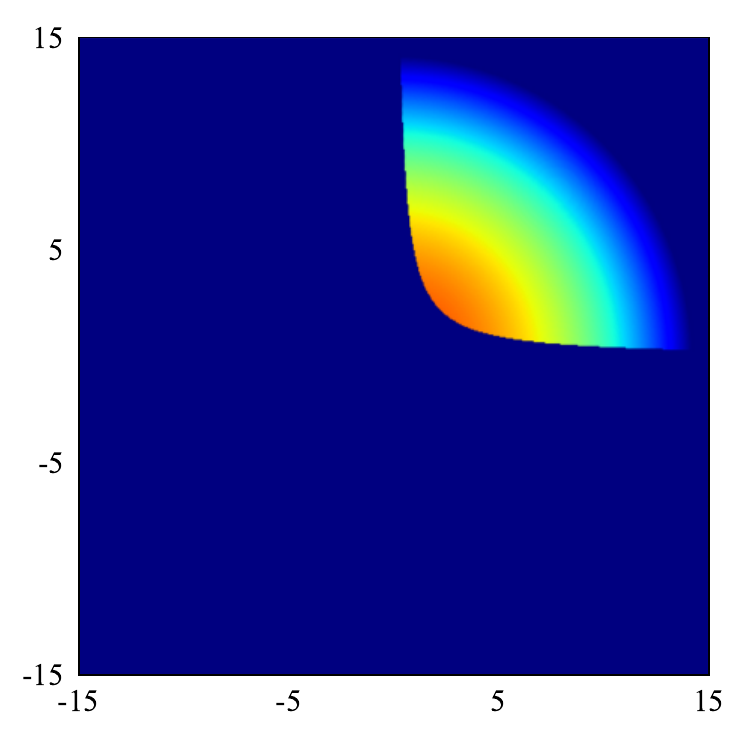}
	\includegraphics[width=0.19\linewidth]{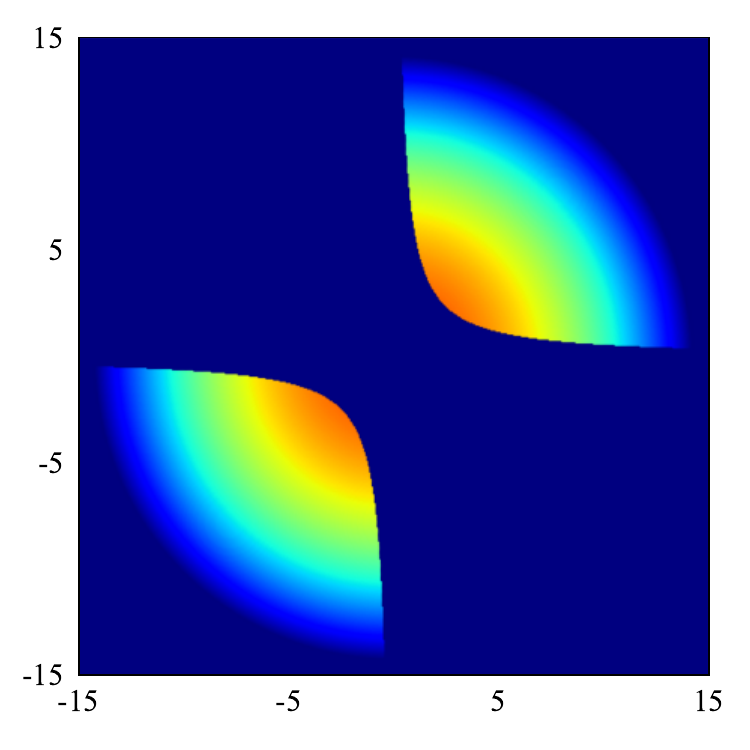}
  \includegraphics[width=0.19\linewidth]{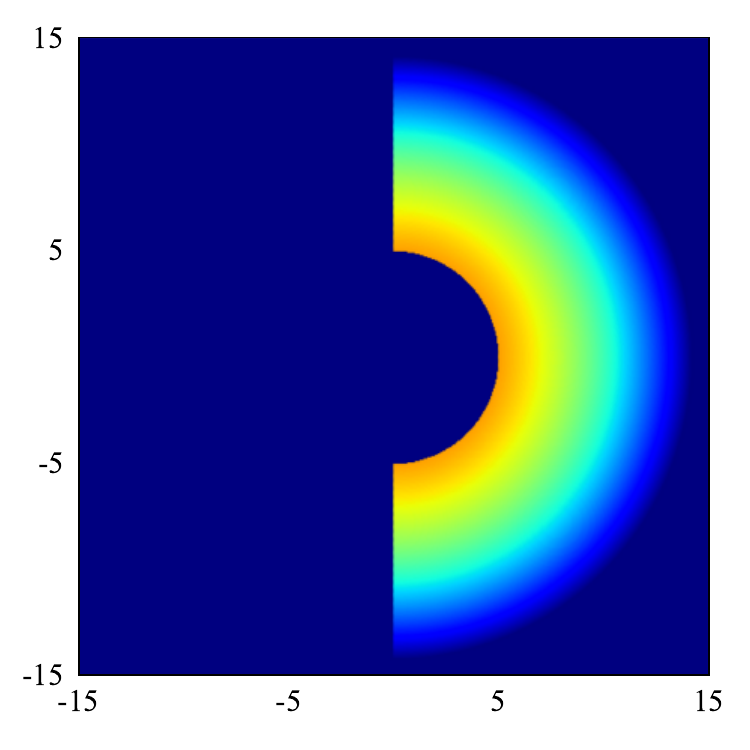}
  \includegraphics[width=0.19\linewidth]{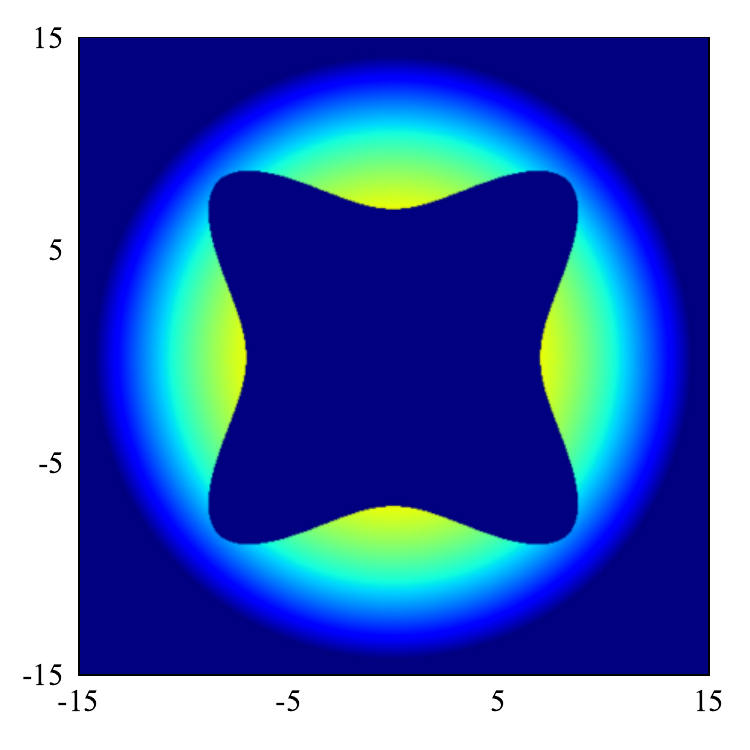}
  \includegraphics[width=0.19\linewidth]{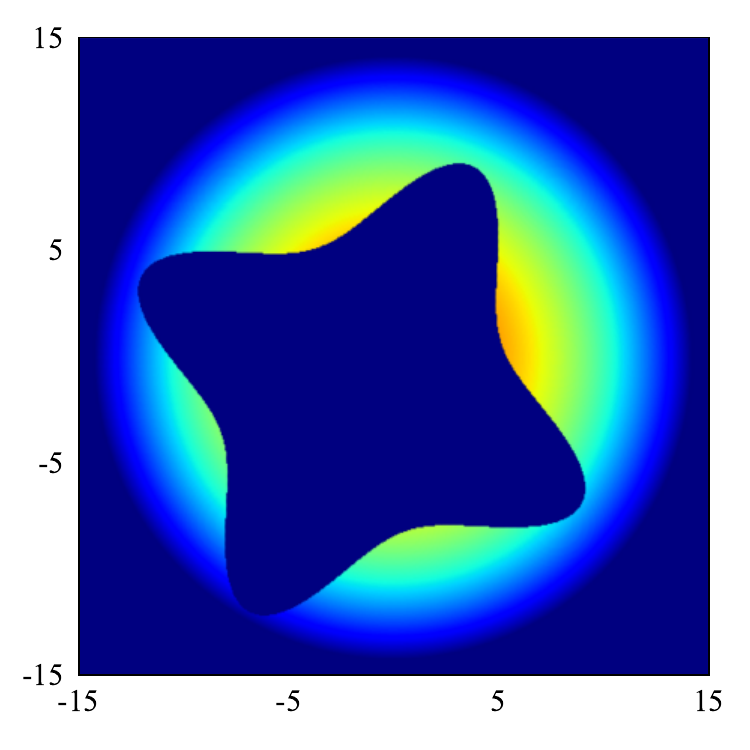}\\
  % \vspace{-0.4in}       
  \vspace{-6pt}
	\includegraphics[width=0.19\linewidth]{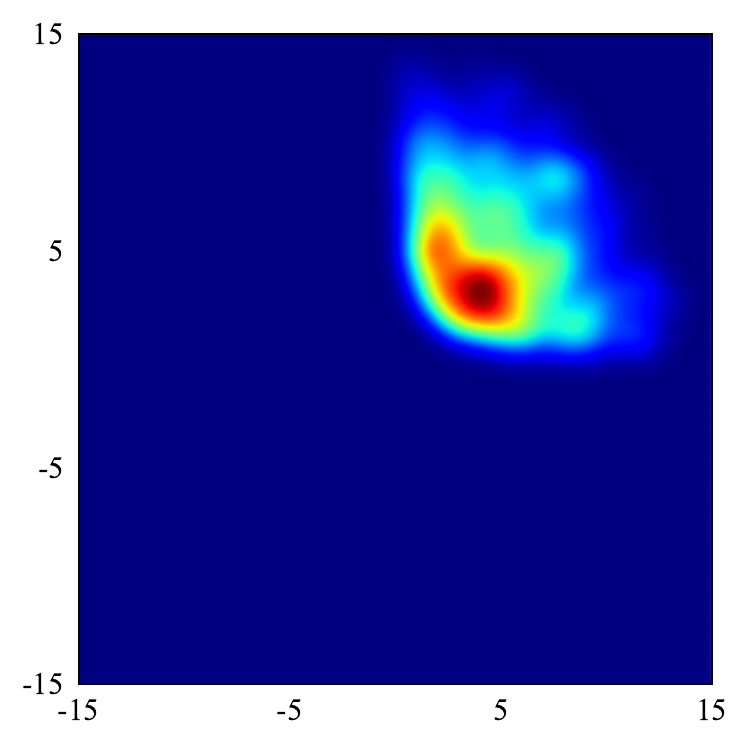}
	\includegraphics[width=0.19\linewidth]{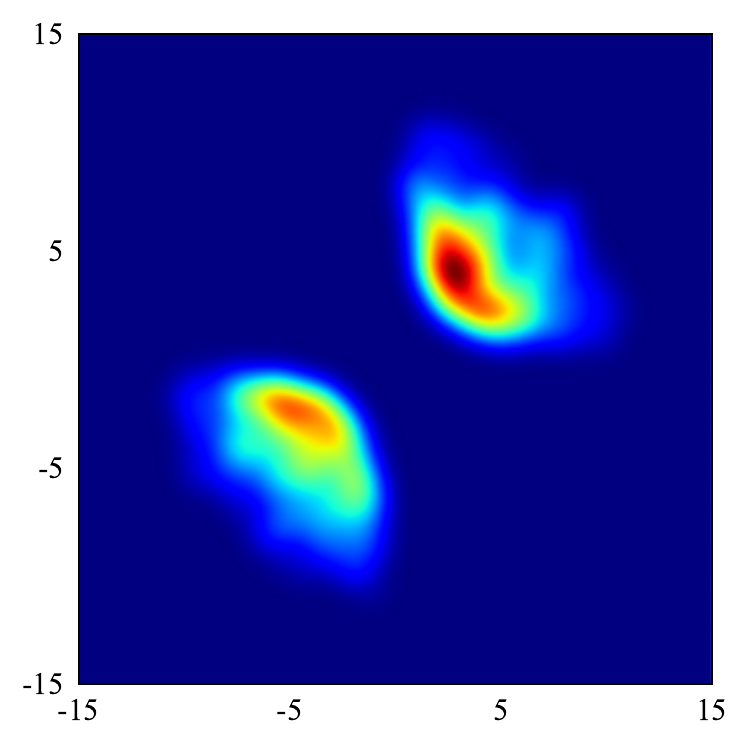}
  \includegraphics[width=0.19\linewidth]{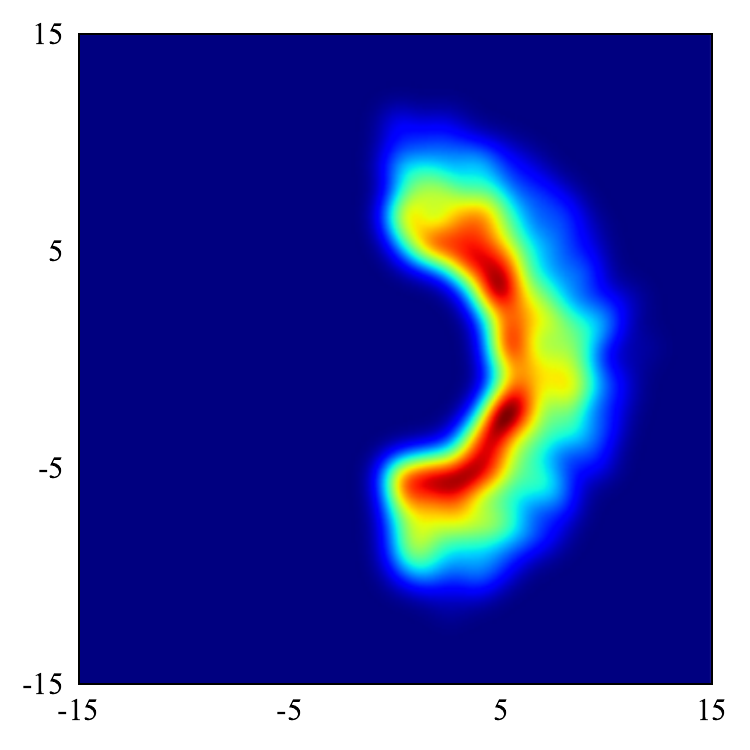}
  \includegraphics[width=0.19\linewidth]{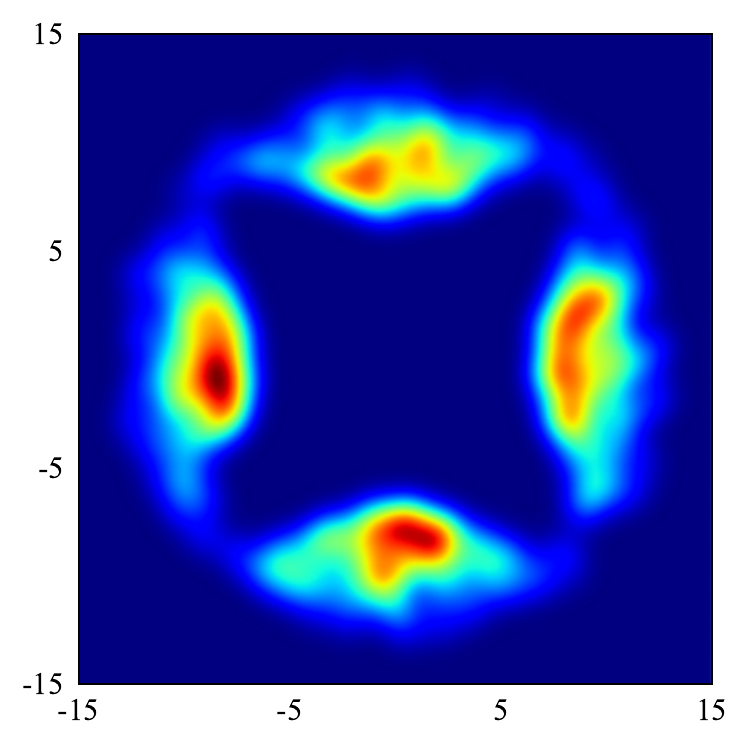}
  \includegraphics[width=0.19\linewidth]{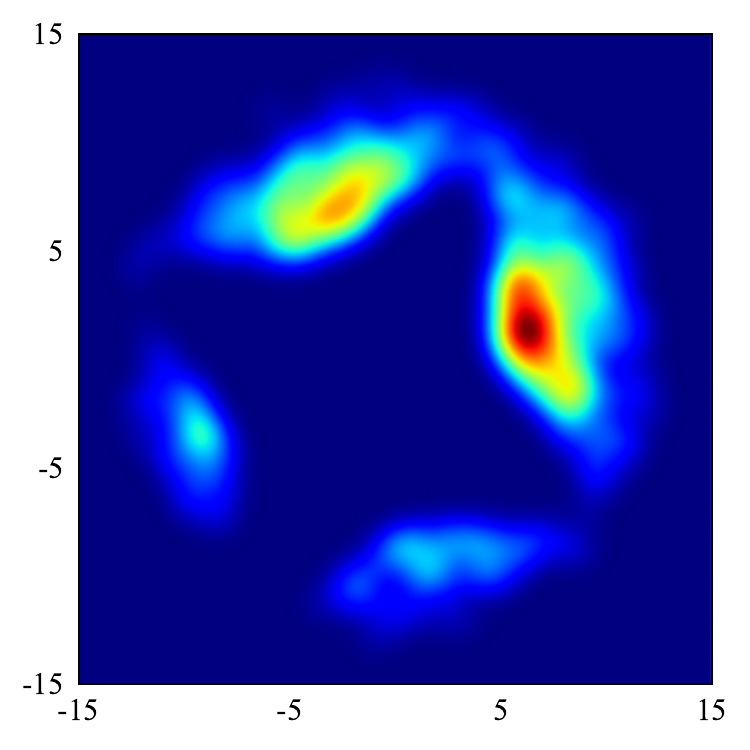}\\
  \vspace{-6pt}
	\includegraphics[width=0.19\linewidth]{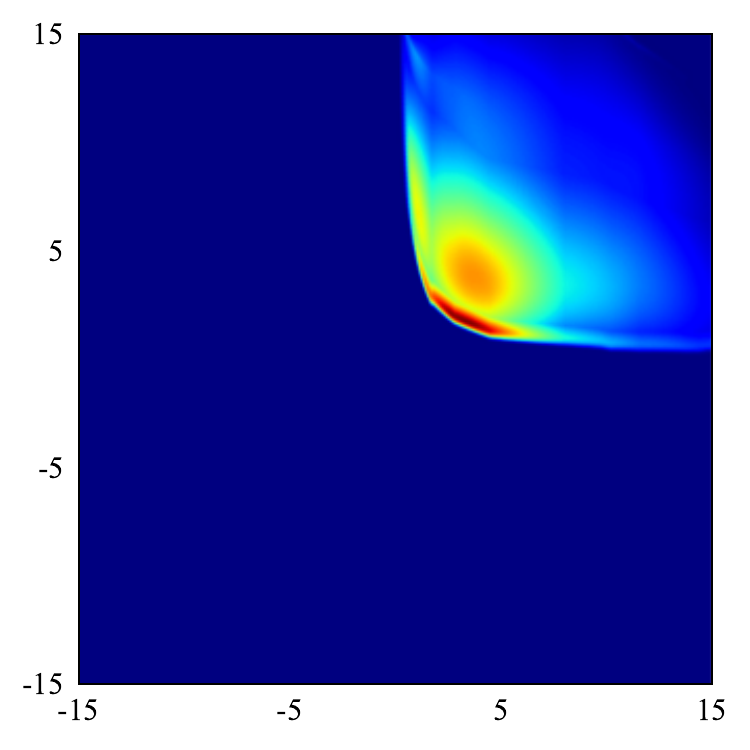}
	\includegraphics[width=0.19\linewidth]{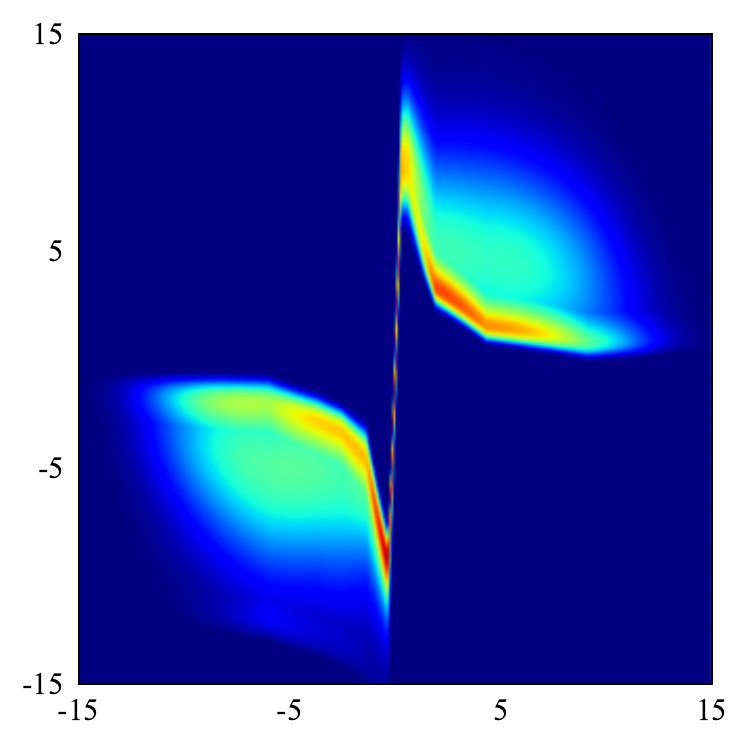}
  \includegraphics[width=0.19\linewidth]{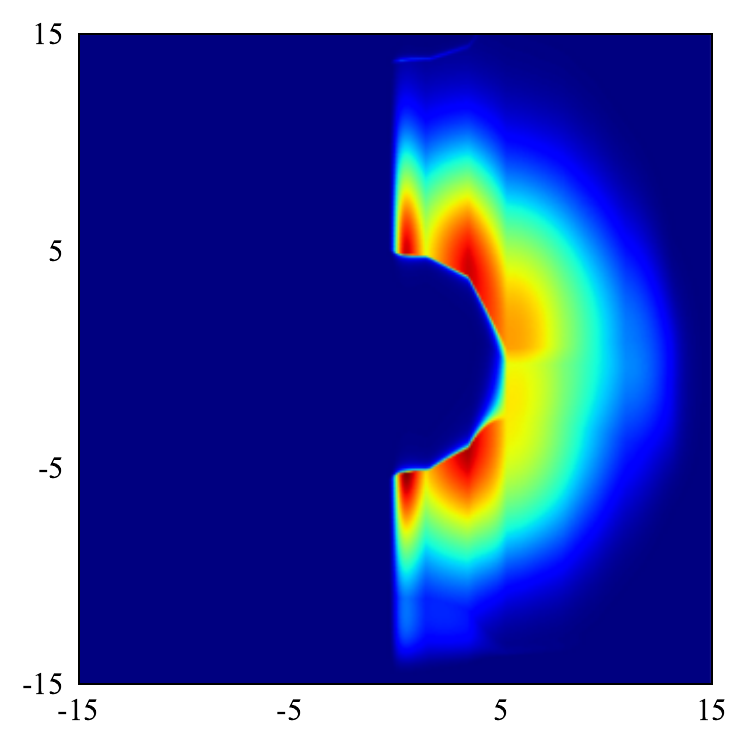}
  \includegraphics[width=0.19\linewidth]{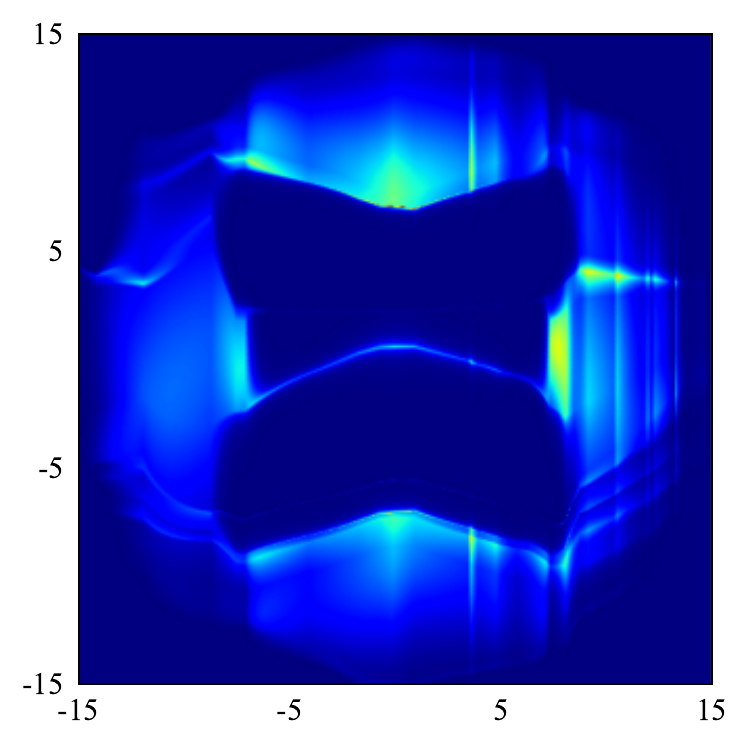}
  \includegraphics[width=0.19\linewidth]{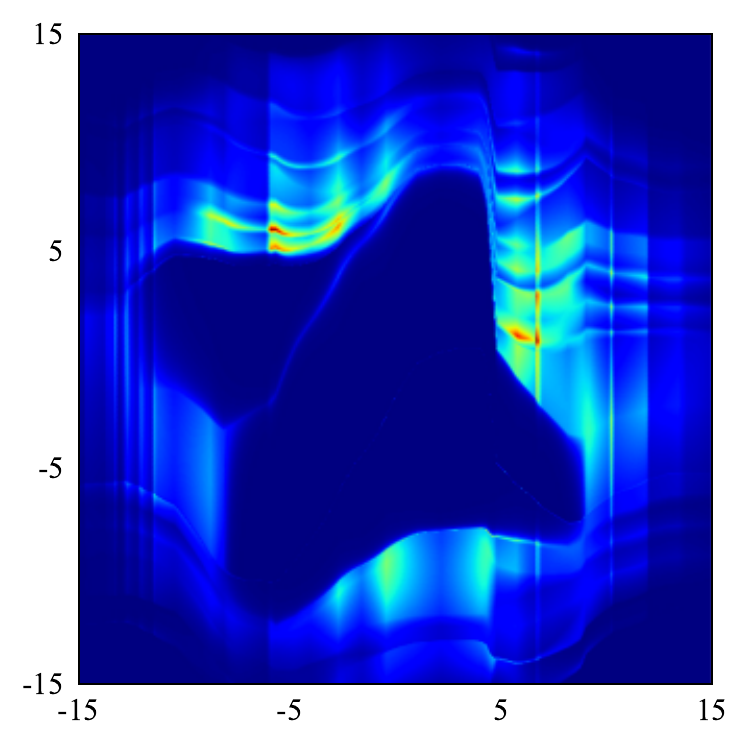}
  \end{minipage}
  \caption{Illustration of \ours in five (columns) 2D toy examples. Top row: the true log failure probability. Middle row: the log failure probability estimated with KDE based on onion sampling with 1000 samples. Bottom row: the log failure probability estimated by NF based on the onion samples.}
  \label{fig:toy}
  \vspace*{-0.2in}
\end{figure*}
\section{Proposed Approach}
% \section{Research methodology}
% In this section, we review the classic NM method and reveal its main issue.
% We then formulation the optimal proposal $q(\x)$ as a variational optimization problem, which reveals that the classic NM method is a simplification solution of the variational problem.
% This naturally lead us fix to improve NM. 
% We then take a further step to seek a general solution to the variational problem, which lead to the key component in solving modern yield estimation problem---optimal manifold.
% % 
% (The optimal manifold also reveals the connection for IS method and the surrogate method) 
% We then propose a novel method based on the latest advance in machine learning, namely, normalization flow to approximate the optimal proposed distribution $q(\x)$.

\ours relies on two key components: onion sampling and NSF, both of which are based on the optimal manifold generalized from NM.
A heads-up example of \ours to approximate five toy failure probability distributions using 1000 samples is shown in Figure \ref{fig:toy} to show the effectiveness. 
\subsection{Optimal Proposal Distribution in IS}
% \Wei{add the variance of IS show the optimal proposal distribution}
From \Eqref{eq:IS}, we can see that the optimal proposal distribution $q(\x)$ is the one that minimizes the approximate variance given by the Delta method, \ie
\begin{equation}
  q^*(\x)= \argmin_{q} \EE_{q} \left[ w^2(\x) \left( I(\x)-\hat{P}_f \right)^2 \right].
\end{equation}
Utilizing Lagrange multiplier rule for calculus of variations, we can show that the optimal proposal distribution is given by
\begin{equation}
  \label{eq:optimal_proposal}
  % q^*(\x) = \frac{p(\x)}{\hat{P}_f} \frac{1}{\sqrt{2 \pi \hat{\sigma}^2}} \exp \left( - \frac{(\x-\hat{\bmu})^2}{2 \hat{\sigma}^2} \right),
  % q^*(\x) \propto p(\x) |I(\x)-\hat{P}_f|,
  q^*(\x) = p(\x)I(\x)/\hat{P}_f.
  % q^*(\x) = \frac{1}{\hat{P}_f} p(\x)I(\x)
  % q^*(\x) \propto p(\x)I(\x),
\end{equation}
If we take a Laplace approximation, we have
\begin{equation}
  % q^*(\x) \approx \frac{ p(\x)I(\x)}{\hat{P}_f} \exp\left(-{(\x-\hat{\bmu})^2} \S^{-1} {(\x-\hat{\bmu})^2} \right),
  q^*(\x) \approx p(\x)I(\x) \exp\left(-{(\x-\hat{\bmu})^T} \S^{-1} {(\x-\hat{\bmu})} \right) / {\hat{P}_f},
\end{equation}
% where 
% \begin{equation}
%   \begin{aligned}
%     \hat{\bmu} &= \argmax_\x \log p(\x)I(\x) \\
%     \S^{-1} &= \nabla^2_{\x} \log p(\x)I(\x) \big|_{\x=\hat{\bmu}}
%   \end{aligned}
% \end{equation}
where 
% $\hat{\bmu} = \argmax_\x \log p(\x)I(\x); \S^{-1} = \nabla^2_{\x} \log p(\x)I(\x) \big|_{\x=\hat{\bmu}}$.
$\hat{\bmu} = \argmax_\x \log p(\x)I(\x); \S^{-1} = \nabla^2_{\x} \log p(\hat{\bmu})I(\hat{\bmu})$.
Notice that $p(\x)$ is just a Gaussian, which is monotonically decreasing in $|\x|$. Thus, $\hat{\bmu}$ is obtained for the smallest $\x$ that satisfies $I(\x)=1$,
exactly the solution in NM of \Eqref{eq:OSV}. 
Furthermore, the Laplace approximation indicates how to design the covariance properly. However, $I(\x)$, in this case, is not a continuous function, and the computation of $\S^{-1}$ is ill-posed. That is probably the reason most previous works use a preset diagonal covariance \cite{MNIS,AIS}, leading to inferior performance due to the ignorance of the variance, which we will discuss later.
% To derive a valid $\S$ under the Laplace approximation, we may relax the hard constraint $I(\x)=1$ using a sigmoid function.

However, the main issue of NM is that it seeks only the closest single failure region and ignores other failure regions, leading to inferior performance. For instance, in \Figref{fig:toy}, NM will only work for the first case as the other four cases have multiple failure regions.
To resolve this issue, the SOTA methods seek a mixture of distributions (\eg vMF \cite{NGAIS}) to approximate the optimal proposal distribution.
However, the number of mixture components is highly problem-dependent, making this approach impractical for real problems.

% \Wei{
% % \Wei{Analysis of other OSV based method}
% As for deriving $\S$ under the Laplace approximation, one way might be relaxing the hard constraint $I(\x)=1$ using a sigmoid function.
% One effective way to improve NM is probably instead introducing a Gamma distribution, which can preserve its main volume at the failure regions it the orientation is done right.
% 
% The biggest issue of NM is that it only seeks closest single failure region without considering its volume and will also ignore other failure regions.
% We show this phernoemona in the \ref{}.
% % 
% To resolve this issue, the SOTA methods seek a mixture of Gaussian other distribution (\eg vMF \cite{NGAIS}) to approximate the optimal proposal distribution.
% However, the number of mixture components is hyperparameter that needs to be tuned according to different problems, making this method impractical for real problems.
% }

\subsection{Optimal Manifold}
% Similarly to Gaussian mixture idea, let introduce a mixture
To enhance NM for more complex problems, let us equip it with an infinite mixture of Gaussian to drastically improve its model capacity
$q(\x) = \sum_{i=1}^M w_i \mathcal{N}(\x-\bmu_i, s\I),$
% \begin{equation}
%   \label{eq: mix normal}
%   q(\x) = \sum_{i=1}^M w_i \mathcal{N}(\x-\bmu_i, s\I),
% \end{equation}
where  $w_i$ is the weight and $\M \rightarrow \infty$.
% Note that we do not need a weight for each Gaussian component because we have an infinite of them and can form any distribution we want.
% 
% $q(\x) = \sum_{i=1}^M \frac{1}{M} \mathcal{N}(\x-\bmu_i, s\I)$.
% \begin{equation}
%   % \label{eq:NM}
%   q(\x) = \sum_{k=1}^K \pi_k \frac{1}{\sqrt{2 \pi \hat{\sigma}^2}} \exp \left( - \frac{|\x-\hat{\bmu}_k|^2}{2 \hat{\sigma}^2} \right),
% \end{equation}
% where $\pi_k$ is the weight of the $k$-th Gaussian component and $\hat{\bmu}_k$ is the mean of the $k$-th Gaussian component.
% To significantly improve capacity for the proposal distribution, we introduce a mixture of Gaussian 
% 
Unlike the Laplace approximation, to avoid ignorance of the variance, we minimize the KL divergence 
% $\KL(q(\x)||q^*(\x))=$, 
% \begin{equation}
%   \label{eq:optimal_manifold}
%   \begin{aligned}
%      \KL(q(\x)||q^*(\x)) & =  \EE_{q(\x)} \left[ \log q(\x) \right] - \EE_{q(\x)} \left[ \log q^*(\x) \right] \\
%      & geq 
%      & \int \sum_{i=1}^M \frac{1}{M} \mathcal{N}(\x-\bmu_i, s\I)  \log \left(p(\x)I(\x)/\hat{P}_f \right) \\
%   \end{aligned}
% \end{equation}
\begin{equation}
  \label{eq:KL}
  \KL(q^*(\x)||q(\x)) =  \EE_{q^*(\x)} \left[ \log q^*(\x) \right] - \EE_{q^*(\x)} \left[ \log q(\x) \right],
    %  & \geq  -\EE_{q^*(\x)} \left[ \log q(\x) \right]. \\
\end{equation}
where $\EE_{q^*(\x)} \left[ \log q^*(\x) \right]$ is a constant of the entropy of the optimal proposal distribution. 
Minimization of the KL divergence is equivalent to maximizing $\EE_{q^*(\x)} \left[ \log q(\x) \right]$, which is the low bound of the optimal solution.
% Substituting the mixture distribution into \Eqref{eq:KL}, we aim to optimize
We now aim to optimize
\begin{small}
\begin{equation}
  \label{eq:optimal_manifold}
  \argmax_{\{\bmu_i, w_i\}_{i=1}^M} \int  p(\x)I(\x)/\hat{P}_f   \log \left(\sum_{i=1}^M w_i \mathcal{N}(\x-\bmu_i, s\I)  \right) \d \x.
\end{equation}
\end{small}
% \begin{equation}
%   \label{eq:optimal_manifold}
%   \begin{aligned}
%     % \argmin \KL(q(\x)||q^*(\x)) = \argmax \EE_{q(\x)}[\q^*(\x)]
%     &\argmax \EE_{q(\x)}[\q^*(\x)]\\
%     = & \argmax_{\{ \bmu_i\}_{i=1}^M} \left( \int \sum_{i=1}^M \frac{1}{M} \mathcal{N}(\x-\bmu_i, s\I)  \log \left(p(\x)I(\x)/\hat{P}_f \right)  \right)\\
%     = & \argmax_{\{ \bmu_i\}_{i=1}^M} \left( p(\x)I(\x)/\hat{P}_f   \log \left(\int \sum_{i=1}^M \frac{1}{M} \mathcal{N}(\x-\bmu_i, s\I) \right)  \right)
%   \end{aligned}
% \end{equation}
% Unlike the Laplace approximation, to avoid ignorance of the variance, we propose to minimize the KL divergence
% $\argmin \KL(q(\x)||q^*(\x))$, 
% instead of matching the maximum probability.
% Because $\EE_{q(\x)}[\q(\x)]$ is a constant, we rewrite the optimization problem as \Wei{wrong?}
% we put forth the following maximization of the proposal distribution $q(\x)$,
% \begin{equation}
%   \label{eq:optimal_manifold}
%   \begin{aligned}
%     % \argmin \KL(q(\x)||q^*(\x)) = \argmax \EE_{q(\x)}[\q^*(\x)]
%     &\argmax \EE_{q(\x)}[\q^*(\x)]\\
%     = & \argmax_{\{ \bmu_i\}_{i=1}^M} \left( \int \sum_{i=1}^M \frac{1}{M} \mathcal{N}(\x-\bmu_i, s\I)  \log \left(p(\x)I(\x)/\hat{P}_f \right)  \right)
%   \end{aligned}
% \end{equation}

The complete solution to \Eqref{eq:optimal_manifold} might seem complicated.
But we can see that to get the maximum, the main volumes of the Gaussian components (corresponding to a large $w_i$) should be placed near the failure boundaries $B(\x)=\partial I(\x) / \partial x \neq 0$. We call this the optimal manifold, which does not necessarily equal $B(\x)$.
In practice, we cannot work with a Gaussian mixture with infinite components. The optimal manifold suggests a suboptimal solution to \Eqref{eq:optimal_manifold} with a finite number of Gaussian components.
% 
% Since $\p(x)I(\x)/P_f$ is just 
% Since $\log \left( \sum_{i=1}^M w_i \mathcal{N}(\x-\bmu_i, s\I) \right)$ is just 
% The complete solution to \Eqref{eq:optimal_manifold} will give us a manifold in the variational parameter located beyond the failure boundaries $B(\x)=\partial I(\x) / \partial x \neq 0$. We call this the optimal manifold, which does not necessary equal to $B(\x)$.
% 
If we set $M=1$, the optimal manifold becomes a variational version of NM, which should provide better performance than the ordinary NM. 
Nevertheless, even the variational NM is difficult to seek because $I(\x)$ is unknown, which really gives credits to NM for its practicality. 

% \Wei{connection to the surrogate-based method}
The crucial point to get the optimal manifold or the optimal proposal distribution is to approximate $I(\x)$.
In the work of \cite{yin2022efficient}, the authors derive a general framework for active learning with optimal proposal samples, which are precisely the points that reduce the uncertainty of $I(\x)$. 
This gives us a direct connection between IS and surrogate methods for that they actually share the same intermediate step, approximating $I(\x)$, in order to efficiently estimate $\hat{P}_f$.
The difference is that most IS methods update the information about $I(\x)$ implicitly by random samples to reveal the unseen $B(\x)$. Because the samples are generated randomly, this procedure is not target orientated and thus is not as efficient, which also comes with the advantage that it will not be trapped in local minima and will always converge.
On the other hand, surrogate methods update the information about $I(\x)$ explicitly by reducing the uncertainty of $I(\x)$. Because $I(\x)$ is approximated with a regression that is highly non-convex to optimize, and the optimization of the proposal points is again highly complex, this procedure will be trapped in local minima and will not always converge. 
% \Wei{we will show some empirical results in the experimental section.}

To combine both advantages of IS and surrogate methods, we can use a complex generative model, says, normalizing flows (NF), to learn from data and approximate $q^*(\x)$, while the computation of $\hat{P}_f$ is still done by IS with samples drawn from the NF. This way, even if the proposal distribution is suboptimal, the IS methods will always converge. In the meanwhile, the NF can use existing samples to implicitly learn the target failure probability. Although NF is a powerful model, it contains a deep neural network (NN) and is thus data-demanding. 
The challenge left is how to generate as many failure points as possible with limited resources while these samples must come from $q^*(\x)$.

% % \Wei{we use samples from NF not building a surrogate}
% % Inspired by the optimal sphere, we propose a spherical pre-sampling method to sample from $p(\x)I(\x)/\hat{P}_f$ and build a initial dataset for NF training.
% % The challenge is how do to give a good initial dataset for NF training.
% % The crucial criteria is that we should have as many as training data as possible; in the same time, this samples must sample from $p(\x)$, which then are less likely to fall into the failure regions.
% % 
% % 
% \Wei{
% Since approximating $I(\x)$ is equivalent to solving the yield Estimation problem, we should avoid directly relying our solution on  $I(\x)$. 

% We approximate the optimal proposal distribution $\q^*(\x)$ 
% The optimal posterior we are seeking is given by
% % \begin{equation}
% %   \label{eq:optimal_posterior}
% %   % q^*(\x) = \argmin_{R, \{ \bmu_i\}_{i=1}^M} \KL(q(\x)||p(\x)I(\x)/\hat{P}_f),
% %   q^*(\x) = \argmin_{R, \{ \bmu_i\}_{i=1}^M} \KL(q^*(\x)||q(\x)),
% % \end{equation}
% \begin{equation}
%   \label{eq:optimal_posterior}
%   \begin{aligned}
%     & q^*(\x) = \argmin_{R, \{ \bmu_i\}_{i=1}^M} \KL(q^*(\x)||q(\x))\\
%     & = \argmax_{R, \{ \bmu_i\}_{i=1}^M} \left( \int \sum_{i=1}^M \frac{1}{M} \mathcal{N}(\x-\bmu_i, s\I)  \log \left(p(\x)I(\x)/\hat{P}_f \right)  \right) \\
%   \end{aligned}
% \end{equation}
% where $q(\x) = \sum_{i=1}^M \frac{1}{M} \mathcal{N}(\x-\bmu_i, s\I), \; \mathrm{s.t.} \; ||\bmu_i||^2=R$.
% }
% \Wei{connection to the surrogate-based method}

% \vspace*{-0.2in}
\subsection{Optimal Hypersphere}
Before we move to NF and the optimal manifold, we will derive the optimal hypersphere and an effective onion sampling to provide training data for NF.
As discussed, the failure boundary $B(\x)$ is unknown, and there is no way to actually solve the optimization in \Eqref{eq:optimal_manifold} in practice.
We thus turn to an easier solution by constraining the Gaussian centroids lying on a hypersphere of radius $r$, that is $\bmu_i^2=r$. We aim to solve 
\begin{small}
\begin{equation}
  \label{eq:optimal_sphere}
  \argmax_{\{\bmu_i, w_i\}_{i=1}^M} \int  p(\x)I(\x)/\hat{P}_f   \log \left(\sum_{i=1}^M w_i \mathcal{N}(\x-\bmu_i, s\I)  \right) \d \x,
\end{equation}
\end{small}
$\mathrm{s.t.} \; ||\bmu_i||^2=r$.
With $M \rightarrow \infty$, solving \eqref{eq:optimal_sphere} is equivalent to optimizing the radius $t$ such that the integral over $I(\x)p(\x)$ is maximized. We call such a hypersphere with radius $r$ the optimal hypersphere.

Certainly, the optimal hypersphere is also intractable due to the absence of $I(\x)$. However, it shows us that the maximum is achieved by a hollow hypersphere, with its radius $r$ being near the failure boundary $B(\x)$ with the main volume.

% Intuitively, it would be ideal to place Gaussian distributions with mean values exactly along the failure boundary $B(\x)=\d I(\x) / \d x \neq 0$.
% This certainly is not possible because, $ I(\x)$ is not known.
% Nevertheless, we can still seek a hypersphere that approximate $B(\x)$.

% To resolve this issue, we seek infinite number of mixture components to approximate the optimal proposal distribution.
% These mixture components are placed along a hypersphere that approximates the failure boundary $B(\x)$, \ie $||\bmu_i||^2=R$.
% % 
% The optimal posterior we are seeking is given by
% \begin{equation}
%   \label{eq:optimal_posterior}
%   % q^*(\x) = \argmin_{R, \{ \bmu_i\}_{i=1}^M} \KL(q(\x)||p(\x)I(\x)/\hat{P}_f),
%   q^*(\x) = \argmin_{R, \{ \bmu_i\}_{i=1}^M} \KL(q^*(\x)||q(\x)),
% \end{equation}
% 
% \subsection{Connections to Surrogate-based Methods}
% \subsection{Algorithm Scheme}
% \subsection{Sphere Slicing Pre-sampling Onion Sampling}

% \begin{figure}
%   \centering
%   \includegraphics[width=0.75\linewidth]{paper_fig/presampling.pdf}
%   \caption{presampling}
%   \label{fig:fig2}
%   \vspace*{-0.2in}
% \end{figure}
% \subsection{Onion Sampling}

% To approximate $q^*(\x)$, the most important step is to sample from $q^*(\x)$, which takes us back to the original problem of sampling from $p(\x)I(\x)/\hat{P}_f$.
% 
The optimal hypersphere suggests that we can sample inside a hollow hypersphere to efficiently generate samples that are both likely to fail ($I(\x)=1)$, and come from the original parameter distribution $p(\x)$. We propose a novel onion sampling inspired by the optimal hypersphere in this section.
% 
% The key element is to avoid sampling from the non-failure regions as possible, which will makes no contribution when training the NF. 
% Simultaneity, these samples should be sample from  $p(\x)$.
% Inspired by the optimal sphere, we can approximately sample from $p(\x)I(\x)/\hat{P}_f$ with efficiency.

% To this end, we propose the onion sampling based on the optimal sphere.
For a hypersphere with radius $r$, we can compute the cumulative distribution function (CDF) \ie $F(r) = \int p(|\x|<r) \mathrm{d} \x$.
% 
% Inspired by the Latin hypercube sampling \cite{mckay1979comparison}, since each slice share the same probability, we genera $J$ samples in each slice.
Following the Latin hypercube sampling (LHS), we divide the domain with $K$ hollow hyperspheres such that
$F(r_k) = \frac{k}{K}$, where $r_k$ is the radius of the $k$-th hypersphere.
The particular value of $r_k$ is easy to compute because the inverse of $F(r)$ can be computed analytically.
If we select each hypersphere with probability $1/K$ and generate $J$ samples inside the hypersphere using a uniform distribution (which will allow us to effectively explore the domain for failure regions), we can reproduce sampling from $p(\x)$ with precision proportional to $1/K$.

As the optimal hypersphere suggests, we should avoid sampling in the center, which is likely to be a non-failure region for the yield problem.
Instead of choosing a hypersphere randomly, we start from the largest hypersphere $r=r_K$.
Inside the sphere, we sample $J$ points uniformly and put them through the SPCIE, compute $I(\x)$, and keep all failure samples.
We also define the failure rate under the uniform sampling for the $k$-hypersphere as ${U}_k$, which gives us an indication when we approach the failure boundary $B(\x)$ where ${U}_k$ will experience a sudden drop.
We repeat this process until ${U}_k$ is below a threshold $\tau$. 
A smaller $\tau$ results in a more accurate sampling but will also increase the computational cost.
This sampling process is like peeling an onion layer by layer, and thus we call it onion sampling, which is summarized in Algorithm \ref{algo1}.

The onion sampling can be further improved for practical use. We discuss some scenarios here. As discussed in the optimal sphere, to have a good approximation of $q^*(\x)$, the key is to have a good matching of the main volume for $q^*(\x)$.
If we have more but a limited budget left for the pre-sampling stage, we can repeat the previous procedure but start from the ending hypersphere near the optimal hypersphere going outward.
If the $K$-th hypersphere is not reached during this process, the total samples might be a bit distorted from $\q^*(\x)$ but still provide a good training set for NF, which will correct such distortion during its update iterations.
If there is more budget, we might exclude the possible non-failure regions, re-divide the domain into $K$ hyperspheres, and then repeat the onion sampling process. This will give us a closer approximation to the optimal hypersphere.

\begin{algorithm}
	\caption{Onion Sampling}
	\begin{algorithmic}[1]  \label{algo1}
		\REQUIRE SPICE-based Indication $I(\x)$, \# of hypersphere $K$, number of \# per hypersphere $J$, threshold $\tau$
    \STATE Divide the domain into $K$ hyperspheres with equally increased cumulative probability
		% \STATE For each hypersphere sample $J$ samples using uniform distribution
    \WHILE {{$U_k > \tau$}}
    \STATE Uniformly generate $J$ points inside $k$ hypersphere
    \STATE Keep samples that fails, $\X=\X \cup \x_{kj}$ for $I(\x_{kj})=1$
    \STATE Compute uniform failure rate ${U}_k$; $k = k - 1$
    \ENDWHILE
		\RETURN Failure sample collections $\X$
	\end{algorithmic} 
\end{algorithm}
\vspace*{-0.15in}
% \begin{algorithm}
% 	\caption{Onion Sampling}
% 	\begin{algorithmic}[1]  \label{algo1}
% 		\REQUIRE Indication $I(\x)$, number of samples $N_{mc}$
% 		\STATE Initial $R_{initial}$, step $k$
% 		\STATE Update record of best $\x^*$ and best iteration $y^*$
% 		% \STATE Initialize surrogate model $\Mcal$ with default PTA  
% 		\FOR {$i = 1$ to $N_{mc}$}
% 		\STATE Update surrogate model $\Mcal$ by maximizing \eqref{MLE}
% 		\STATE Sample a netlist $\bxi$ from $\mathcal{Q}$
% 		\STATE Optimize acquisition function and get optimal $\x$
% 		% \IF{sd} 
% 		% \STATE sd
% 		% \ENDIF s
% 		% \STATE Optimize the acquisition function \eqref{EIfunc}, \eqref{eqUCB}, or \eqref{eq_MES} given $\bxi_i$ and get candidate $\x^*$
% 		% \STATE Inquire best candidates for $\bxi_i$
% 		\ENDFOR
% 		\RETURN Monte-Carlo analysis for $\mathcal{Q}$
% 	\end{algorithmic} 
% \end{algorithm}

% we can sample from $F(\x')$ to obtain a set of samples $\{ \x_i\}_{i=1}^M$.

\cmt{
\Wei{Presampling (from Yanfang)}
The purpose of pre-sampling is to obtain failure samples to construct the intermediate sampling distribution. 
% In order to speed up the yield calculation, the importance sampling method needs to be adjusted to reduce the variance. The variance is derived from the mismatch between the intermediate and most sampled distributions. Therefore, finding a distribution that approximates
% \begin{equation}
%     g^{*}(\x)=\frac{f(\x)I(\x)}{\int f(\x)I(\x)d\x}
% \end{equation}
% as the intermediate sampling distribution is very important to improve the yield calculation speed. From the expression for the optimal sampling distribution, we see that the denominator is a constant. 
% Therefore, the middle sampling distribution should be as proportional as possible to $f(\x)I(\x)$.
% where $f(X)$ represents the input distribution and $I(\x)$ is the indicator function when $\x=1$ for failure and $\x=0$ for non-failure. 
For the failure samples, the indicator function is constant , so an efficient intermediate sampling distribution needs to cover the failure points as much as possible and be proportional to $f(\x)$. 
However, if $f(\x)$ is used directly for sampling, the probability of collecting failure points is very low, and if $f(\x)$ is not used for sampling, the distribution constructed will not match the optimal sampling distribution. 
Therefore, we find a pre-sampling method which can not only collect the failure sample points, but also ensure that the distribution of the failure sample points is approximately proportional to $f(\x)$. 
% \begin{figure}
%     \centering
%     \includegraphics[width=0.65\linewidth]{paper_fig/pre2.jpeg}
%     \caption{The sampling distribution of the proposed method.}
%     \label{fig:fig1}
% \end{figure}

% First, we assume that the d-dimensional $\x=[x^{(1)},x^{(2)},\cdots,x^{(d)}]^T \in \x$  are independent and identically distributed from the standard normal distribution $\mathcal{N}(0,1)$, Then the probability density function of the variable $\x$ is 
% \begin{equation}
% \begin{split}
%     P_x= \prod \limits_{i=1}^{d}\frac{1}{\sqrt{2 \pi}} \exp\{ {- \frac{(x^{(i)})^2}{2}}\} \\
%     =(\frac{1}{\sqrt{2 \pi}})^{d}\exp\{ {- \frac{\sum_{i=1}^{d}(x^{(i)})^2}{2}}\}
% \end{split}
% \end{equation}
From (7), we can see that the probability density of the variable X is only related to the distance $\sum_{i=1}^{d}(x^{(i)})^2$ of $\x$ from the origin, independent of the direction of the variable $\x$. 
Therefore, X is uniformly distributed when projected onto a hyperhypersphere\Wei{(hypersphere?)} of the same radius. Multidimensional independent normal distribution sampling, from the point of view of hypersphere, is equivalent to sampling the radius first, and then sampling uniformly on the radius.

When calculating the yield, $\x$ refers to the process parameter, and the larger the distance $ \sum_{i=1}^{d}(x^{(i)})^2 $ from $\x$ to the origin is, the larger the fluctuation of $\x$ is. When $\x$ is close to the base point, the probability of circuit failure will not increase, and when $\x$ is far from the base point, the probability of circuit failure will increase. That is, the larger the radius of the hypersphere, the greater the probability of circuit failure. Because the calculation method of importance weight is

The flow of our pre-sampling method is as follows:
According to the above analysis, our pre-sampling method has the following advantages: 
(1) Is closer to the optimal sampling distribution $g^{*}(\x)$. The sampling of $f(\x)$ on a hyperhypersphere of the same radius behaves as a uniform fraction Cloth, which is exactly the same as the pre-sampling method. 
(2) Sample efficiency is high. The hyperhypersphere interpolation method does not need to sample in the whole space, but only on the hypersphere. As many failure points as possible can be collected on a hyperhypersphere with a larger radius. 
(3) Is simple to implement. 
}

% \subsection{Optimal Manifold}
% Solving \eqref{eq:optimal_posterior} is non-trivial because $I(\x)$ is not known. Further more, the optimal hypersphere also has its issues: (1) the optimal hypersphere consist infinite Gaussian mixture components, which have approximately half volume outside the failure regions, making it less efficient when it comes to sampling from $q(\x)$; (2) the realistic failure boundary is not a hypersphere but a complex manifold, which will lead to a inferior solution (but not incorrect).
% % 
% To resolve these issues, we seek a manifold that approximates $q^*(\x)$ using normalization flow (NF), which is expect to approximate any complex distribution using the power of deep learning and massive parallel computing through GPU. Most Importantly, this method can automatically update itself as more samples are collected to reveal the true failure boundary (as is done in a surrogate-based method).
% \Wei{HD issues} Another advantage to use NF is tha the underlying dimension of the manifold is automatically handled thought the NF distribution, where the contribution of each input in $\x$ is automatically weighted by the NF model \cite{}.
\vspace{-0.05in}
\subsection{Normalizing Flows For Optimal Proposal Distribution}
\label{sec:flow}
The onion sampling is a simple yet effective method for sampling approximately from the optimal proposal distribution $q^{*}(\x)$ relying on the optimal hypersphere, which is a suboptimal solution based on the optimal manifold.
With samples from onion sampling, we now harness the power of modern deep learning and massively parallel computing to approximate the optimal manifold and further improve our performance. 
More specifically, we implement an NF with Neural Spline Flows (NSF) \cite{durkan2019neural}.

% However, the underlying assumption is that the optimal manifold is a sphere, which leads to inferior perform.
% Can we push the frontier further?
% The answer is yes and we can do this using normalizing flows, a method that learns from samples, as active learning is done in a surrogate-based method, and is capable to handle highly complex manifold and approximate any complex distribution by harnessing the power of deep learning.
% However, the sampling method is not suitable for high-dimensional space.
% 
% To well approximate $q^*(\x)$ by harnessing the power of modern deep learning and massive parallel computing, 

NF is a class of generative models that can approximate any complex distributions by transforming a simple base distribution using a series of invertible transformations.
In our case, the base distribution is naturally the standard normal $p(\x)$ whereas the target distribution $q^*(\x)$.
Assume a mapping $\x = \g(\z) $, where $\g:R^{D} \rightarrow R^{D}$ is a bijective function, and $\z \in R^{D}$ is a random vector drawn from a normal distribution $p(\z)$ (which distinguishes itself form $p(\x)$).
The PDF $q(\x)$ can be expressed using the change of variables formula:
% \begin{equation}
%    q(\x)= p(\z)|\operatorname{det}(\frac{\partial \z}{\partial \x})| 
%    = p(\z)|\operatorname{det}(\frac{\partial g}{\partial \x})|
% \end{equation}
% where $g$ is the inverse of $f$, and $ \partial \g  /{ \partial \x} $ is the Jacobian matrix of $\g$.
\begin{equation}
  q(\x)= p( \h(\x) ) |\operatorname{det} D \h(\x)| 
  = p(\z)|\operatorname{det}  D \g( h(\x)) |^{-1},
\end{equation}
where $\h(\cdot)$ is the inverse function of $\g(\cdot)$; $D \h(\x) = \frac{\partial \h(\x)}{\partial \x}$ is the Jacobian matrix of $\h(\x)$; likewise, $D \g(\z) = \frac{\partial \g(\z)}{\partial \z}$ is the Jacobian matrix of $\g(\z)$. Sampling from $q(\x)$ is equivalent to sampling from $p(\z)$ and then applying mapping $\x = f(\z )$.
The key to delivering a close approximation to $q^*(\x)$ is to choose a proper mapping $\g(\cdot)$, which admits an efficient inversion and Jacobian matrix computation. 

After many trial tests with different models (including autoregressive flow, affine coupling flow, planar flows, etc.), we found NSF \cite{durkan2019neural} works the best for the yield problem.
%  for with we give a brief description; for more details, we refer the reader to .
In NSF, $\z$ is divided into two parts, $\z = (\z^{(1:d)}, \z^{(d+1:D)})^T$, and
% \begin{equation}
%   \z^{(1)} = \z^{(1)}, \quad \z^{(2)} = \z^{(2)} + \sum_{i=1}^{d} \alpha_{i}(\z^{(1)}) \z^{(1)}_{i}
% \end{equation}
% \begin{equation}
%   \x^{(1)} = \g_\phi(\z^{(1)}|\bphi), \quad \x^{(2)} = \g_\theta(\z^{(2)}| \btheta(\z^{(1)}))
% \end{equation}
\begin{equation}
  \x^{(1:d)} = \g_\phi(\z^{(1:d)}), \quad \x^{(d+1:D)} = \g_\theta(\z^{(d+1:D)}),
\end{equation}
where 
% $\x^{(1:d)}$ indicates dimension from 1 to $d$, and
$\btheta(\z^{(1:d)})$ is a NN that takes $\z^{(1:d)}$ as input and outputs the model parameters for a spline mapping $ \g_\theta(\z^{(d+1:D)})$.
% $ \g_\theta(\cdot)$ is an elementwise rational spline function that is comprised as the ratio of two polynomials segments
$ \g_\theta(\cdot)$ is a monotonic rational-quadratic spline whose each bin is defined to be the ratio of two
rational-quadratic polynomials \cite{durkan2019neural}.
Training the NF is straightforward by maximum likelihood estimation (MLE),
$\Lcal= \sum_{i=1}^N \log q(\x_i)$, using stochastic gradient descent.
The gradient of the log-likelihood is easily computed via modern automatic differentiation tools based on chain rules.
% 
% \Wei{Avoid collapsing:
% To avoid falling into mode collapse, in the loss function, we add entropy to make the samplings more rich which can accelerate the convergence rate of failure probability, thus the loss function is as follows:
% \begin{equation}
% L=\log p_{\X}(\x ; \theta)-\int P(\x)logP(\x)d\x
% \end{equation}
% where $p_{\X}$ is the PDF of $\x$ and $P(\x)$ is the CDF of $\x$.
% }
% 
% \Wei{HD issues}

NSF offers an excellent combination of functional flexibility whilst maintaining a numerically stable inverse that is of the same computational and space complexities as the forward operation. This element-wise transform permits the accurate representation of complex multivariate distributions.
Another advantage of NSF is that it can deal with high-dimensional problems through the flow by identifying the most important dimensions and then focusing on them.

%%%%%%%%%%%%%%%%%%%%%%%%%%%%%%%%%
\cmt{
Normalizing Flows are a family of methods for constructing flexible learnable probability distributions that allow
both density estimation and sampling, often with neural networks, which allow us to surpass the limitations of simple parametric forms. If  our interest is to estimate the density function $p_{\X}$ of a random vector $\X \in R^{D}  $, then normalizing flows assume $\X = f(\Z) $, where $f:R^{D} \rightarrow R^{D}$ is a bijective function, and $Z \in R^{D}  $ is a random vector with a simple density function $p_{\Z}$. The probability density function  $p_{\X}$ can be expressed using the change of variables formula:
\begin{equation}
\begin{split}
 p_{\X}(\x)= p_{\boldsymbol{Z}}(\z)|\operatorname{det}(\frac{\partial \z}{\partial \x})| \\
 = p_{\Z}(\z)|\operatorname{det}(\frac{\partial g}{\partial \x})|
\end{split}
\end{equation}
where $g$ is the inverse of $f$ which is easy to evaluate and computing the Jacobian determinant takes $\Ocal(D)$ time and $ \partial \g  /{ \partial \x} $ is the Jacobian matrix of $\g$.  Sampling from $p_{X}$ can be done by first drawing a sample from the simple distribution $z\sim p_{\Z}$, and then apply the bijection $\X = f(\Z ) $.

\begin{figure}
    \centering
    \includegraphics[width=1\linewidth]{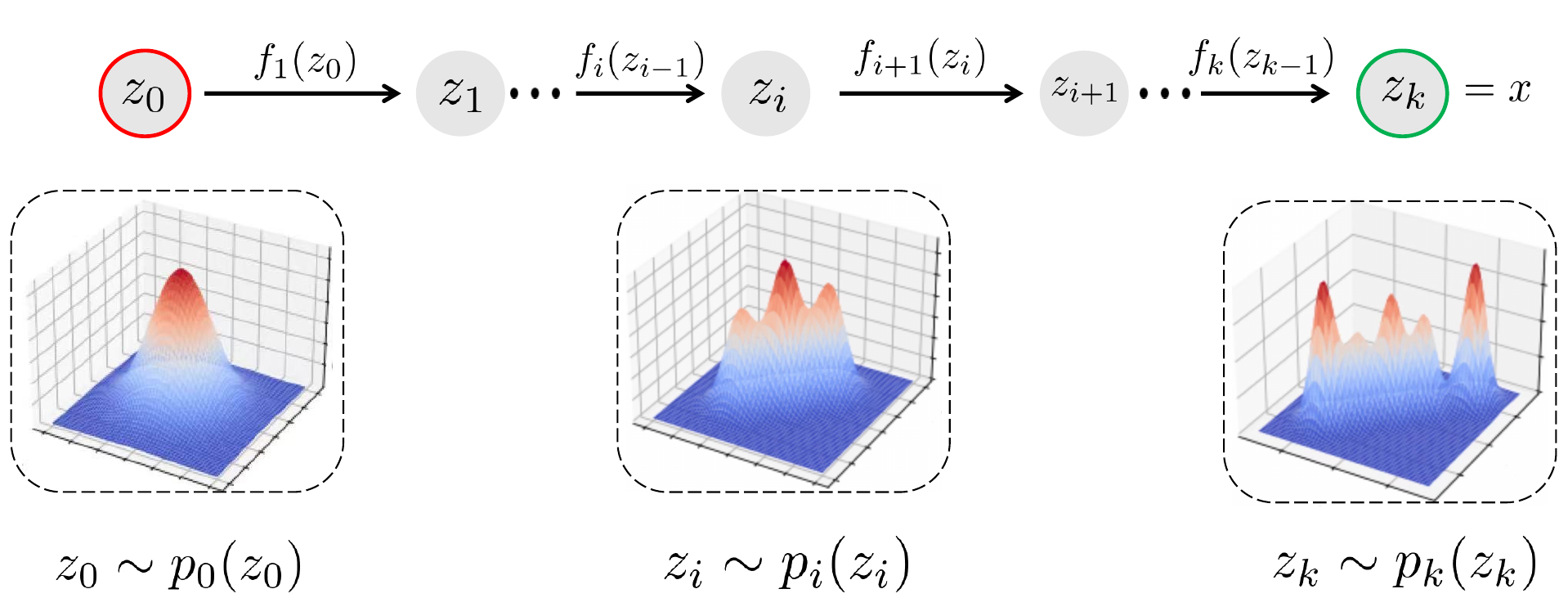}
    \caption{Normalizing Flows.}
    \label{fig:fig2}
\end{figure}

About the bijection, a coupling transform divides the input variable into two parts and applies an element-wise bijection to the section half whose parameters are a function of the first. Optionally, an element-wise bijection is also applied to the first half. Dividing the inputs at $d$ , the transform is,

\begin{equation}
\begin{aligned}
\x_{1: d} &=f_{\theta}\left(\z_{1: d}\right) \\
\x_{(d+1): D} &=h_{\phi}\left(\z_{(d+1): D} ; \z_{1: d}\right)
\end{aligned}
\end{equation}

where $\z_{1: d}$ represents the first $d$ elements of the inputs, $f_{\theta}$  is either the identity function or an elementwise bijection parameters $\theta$ , and $h_{\phi}$  is an element-wise bijection whose parameters are a function of $\z_{1: d}$.

Difference choices for $f$  and $h$ form different types of coupling transforms. When both are monotonic rational splines, the transform is the spline coupling layer of Neural Spline Flow. 

The rational splines are functions that are comprised of segments that are the ratio of two polynomials. For instance, for the $d$-th dimension and the $k-th$ segment on the spline, the function will take the form
\begin{equation}
    x_{d}=\frac{\alpha^{(k)}\left(z_{d}\right)}{\beta^{(k)}\left(z_{d}\right)}
\end{equation}
where $\alpha^{(k)}$ and $\beta^{(k)}$ are two polynomials of order $d$. For $d=1$, we say that the spline is linear, and for $d=2$ , quadratic. The spline is constructed on the specified bounding box, $[-K, K] \times[-K, K]$ , with the identity function used elsewhere. Rational splines offer an excellent combination of functional flexibility whilst maintaining a numerically stable inverse that is of the same computational and space complexities as the forward operation. This element-wise transform permits the accurate representation of complex multivariate distributions.
}

% As shown above, when we combine a sequence of coupling layers sandwiched between random permutations so we introduce dependencies between all dimensions, we can model complex multivariate distributions.

%  Via the change of variables formula the probability density function (PDF) of the flow given a data point can be written as
% \begin{equation}
% \begin{split}
% \log p_{\X}(\x)=\log p_{\Z}(\z)+\log |\operatorname{det}(\partial \g / \partial \x)| 
% \end{split}
% \end{equation}

% This model, parameterized by the weights of the scaling and translation neural networks $\theta$, is then trained via stochastic gradient descent (SGD) on training data points where we maximize the log likelihood function given by
% \begin{equation}
% L=\log p_{\X}(\x ; \theta)
% \end{equation}

\def\Figref#1{Fig.~\ref{#1}}
\def\Tabref#1{Table~\ref{#1}}

\begin{figure}
  \centering
  \includegraphics[width=0.75\linewidth]{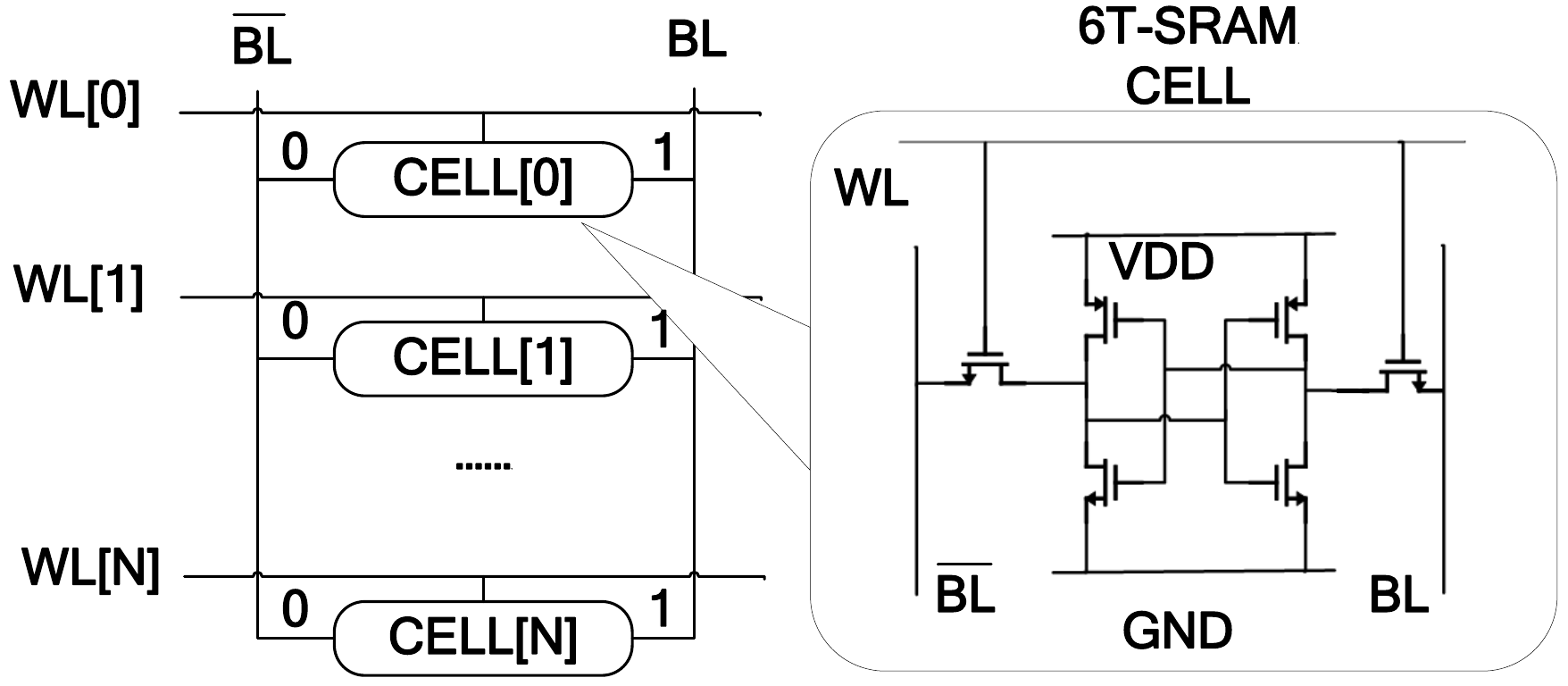}
  \vspace*{-0.1in}
  \caption{The structure of SRAM column circuit}
  \label{SRAM_column}
  \vspace*{-0.2in}
\end{figure}
\section{Experimental Results}
We firstly examine \ours in five toy examples with different artificial failure boundaries (\eg open boundaries, multiple failure regions, and non-centered regions) with their ground-truth log failure probability (LFP) in \Figref{fig:toy}. Onion sampling with 1000 samples and estimated LFP using kernel density estimator (KDE) with a {bandwidth} of 0.75 are shown in the second row. We can see that onion sampling is efficient but also overestimate the LFP. NSF for the estimated LFP is shown in the third row, which shows significant reductions in the overestimation of onion sampling.

We assess \ours on challenging high-dimensional benchmark circuits, namely, three SRAM column circuits with 108, 569, and 1093 variation parameters, respectively. The circuits are synthesized using the Synopsys Design Compiler and the Cadence Virtuoso design tools.
We set the target failure rate of the circuits at approximately $10^{-5}$ to highlight the challenge of the yield estimation problem.
We implement MC as the golden standard to estimate the ground-truth yields.
% The ground truth failure rate is obtained by using MC.  
To show the accuracy and efficiency of \ours, we also implement {the} SOTA IS methods, including Minimized Norm Importance Sampling (MNIS) \cite{MNIS}, Hyperspherical Clustering and Sampling (HSCS) \cite{HSCS}, Adaptive Importance Sampling (AIS) \cite{AIS}, Adaptive Clustering and Sampling (ACS) \cite{ACS}, and surrogate-based methods, including Low-Rank Tensor Approximation (LRTA) \cite{LRTA} and Absolute Shrinkage Deep Kernel learning (ASDK) \cite{ASDK}, as comparison methods.

% and then compare \ours with SOTA IS method, including Mnimized Norm Importance Sampling (MNIS) \cite{MNIS}, Hyperspherical Clustering and Sampling (HSCS) \cite{HSCS}, Adaptive Importance Sampling (AIS) \cite{AIS}, Adaptive Clustering and Sampling (ACS) \cite{ACS}, and surrogate-based methods, including Low-Rank Tensor Approximation (LRTA) \cite{LRTA} and Absolute Shrinkage Deep Kernel learning (ASDK) \cite{ASDK}.
% We assess \ours and other SOTA IS- and surrogate-based yield estimation methods to estimate the failure rate and compare their performance.
% The compared competitors includes Mnimized Norm Importance Sampling (MNIS) \cite{MNIS}, Hyperspherical Clustering and Sampling (HSCS) \cite{HSCS}, Adaptive Importance Sampling (AIS) \cite{AIS}, Adaptive Clustering and Sampling (ACS) \cite{ACS}, Low-Rank Tensor Approximation (LRTA) \cite{LRTA} and Absolute Shrinkage Deep Kernel learning (ASDK) \cite{ASDK}. 
% 
% In this section, we also implement ablation experiments to prove that the proposed pre-sampling method does contribute to the performance.   

% In order to determine when to stop the yield estimation algorithm, we choose
The figure of Merit (FOM) $\rho= \mathrm{std}(P_f) / P_f$ (where $\mathrm{std}(P_f)$ is the stand deviation of estimated yield) is used as the stopping criterion for all methods with $\rho=0.1$ (indicating at least 90\% accurate with 90\% confidence interval) as in many previous works, \eg \cite{MNIS, HSCS, AMSV}. 
For the 108-dimensional case, we use a 4-layer multi-layer perception (MLP), each with 432 hidden units; for the high-dimensional 569 and 1093 cases, we use a 7-layer MLP with 600 hidden units for each layer. ReLu activation is used. The optimization is done with Adam with 500 epochs. The baseline methods use {(default)} setting as suggested in their papers.
All experiments are performed on a Linux system with AMD 5950x and 32GB RAM.
 
 % \help{experimental setting: dense network}

% to indicates a estimation of 
% to terminated the iteratio
% we stop the process of yield estimation and declare that the estimated $P_f$ is 

\begin{table*}
  \centering
      \tabcolsep=5pt
      \caption{Numerical results on three SRAM column circuits}
      \label{AllCaseTable}
      \vspace*{-0.1in}
      \begin{tabular}{l|cccc|cccc|cccc}
        \toprule
        & \multicolumn{4}{|c|}{108-dimensional case} & \multicolumn{4}{|c|}{569-dimensional case}  & \multicolumn{4}{|c}{1093-dimensional case} \\
        \midrule
         Method & Fail. prob. & Rel. error & \# of sim. & Speedup & Fail. prob. & Rel. error & \# of sim. & Speedup & Fail. prob. & Rel. error & \# of sim. & Speedup\\
        \midrule
        MC   & 5.01e-5 &   -   & 699000 & 1x & 2.50e-5 &   -   & 931000 & 1x & 4.80e-5 &   -   & 1189000 & 1x \\
        MNIS & 4.15e-5 & 17.07\% & 47500 & 14.72x & 2.07e-5 & 17.33\% & 59000  & 15.78x & 4.21e-5 & 12.32\% & 81000 & 14.68x\\
        HSCS & 4.84e-5 & 3.36\% & 26500 & 26.38x & 2.86e-5 & 14.27\% & 46500  & 20.02x & 4.30e-5 & 10.47\% & 66000 & 18.02x\\
        AIS  & 4.75e-5 & 5.21\% & 12300 & 56.83x & 2.38e-5 & 4.99\%  & 25700  & 36.23x & 4.43e-5 & 7.75\% & 38000 & 31.29x\\
        ACS  & 5.68e-5 & 13.40\% & 10400 & 67.21x & 2.73e-5 & 9.19\%  & 22500  & 41.38x & 4.42e-5 & 7.83\% & 30400 & 39.11x\\
        LRTA & 4.50e-5 & 10.18\% & 13000 & 53.77x & 2.26e-5 & 9.60\%  & 18500  & 50.32x & 5.25e-5 & 9.38\% & 24000 & 49.54x\\
        ASDK & 4.5e-5 & 10.18\% & 9200 & 75.98x & 2.30e-5 & 8.00\% & 11800 & 78.90x & 6.10e-5 & 27.08\% & 14550 & 81.72x\\
        Proposed & 5.02e-5 & \bf{0.21}\% & \bf{5300} & \bf{131.89x} & 2.49e-5 & \bf{0.25}\% & \bf{3400} & \bf{273.82x} & 4.67e-5 & \bf{2.71}\% & \bf{6400} & \bf{185.78x}\\
      \bottomrule
    \end{tabular}
    \vspace*{-0.1in}
\end{table*}
\subsection{108-Dimensional SRAM Column Circuit} \label{108dim_exp}
An SRAM array is a typical type of random-access memory and uses flip-flop{s} to store data. The SRAM array and cell are shown in \Figref{SRAM_column}, where WL is the word line, and BL is the bit line; two cross-connected inventors composed of four transistors are used for storing data, whereas the other two transistors work as control switches for data transmission.
Each transistor contains three variational parameters.
An 8-bit SRAM array is composed of eight cells, resulting in 108 variation parameters. We choose the delay time of read/write of the SRAM as the output performance metric $\y$ of the circuit. 
%  Our experiment is implemented on a SRAM column circuit with And 
% We compare different methods (MC, MNIS, HSCS, AIS, ACS, LRTA, ASDK, proposed) in accuracy and efficiency.

\begin{figure}[]
  \centering
	\includegraphics[width=1\linewidth]{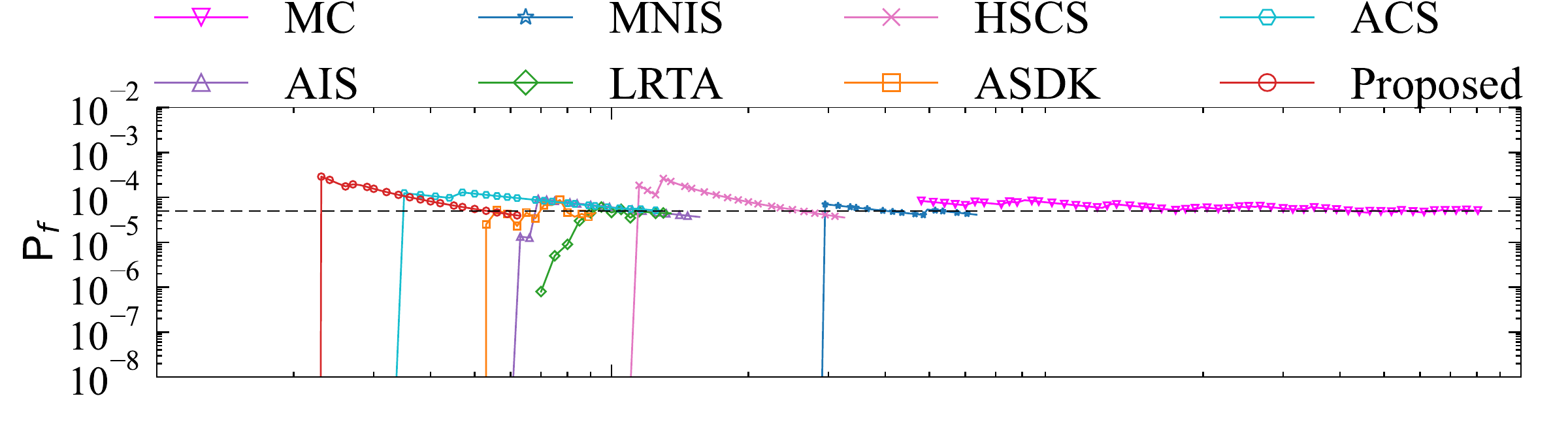}\\
	\vspace{-0.1in}
	\includegraphics[width=1\linewidth]{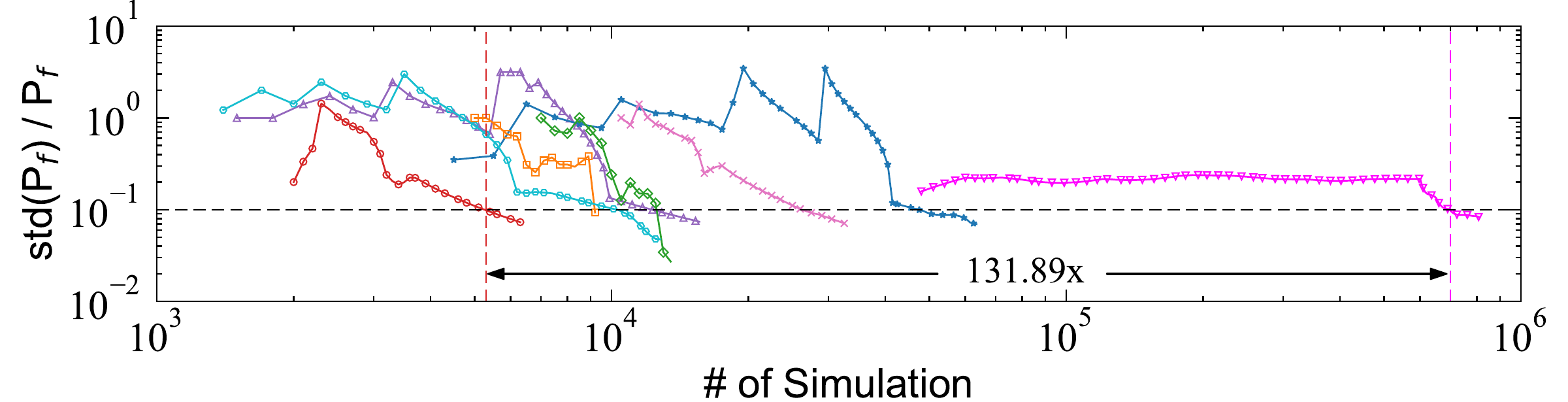}\\
  \vspace{-0.1in}
  \caption{$P_f$ and FOM on 108-dimensional SRAM column}
  \label{case3_exp}
  \vspace{-0.1in}
\end{figure}

% In terms of accuracy, 
We show the failure probability estimation and FOM in \Figref{case3_exp}, and the numerical results are concluded in \Tabref{AllCaseTable}.
The advantage of \ours is obvious by giving a 131.89x speedup compared to MC, more than twice faster {than} the second-best method AIS. Except for being the fastest, \ours also achieves the best accuracy among all methods with a relative error of 0.21\%, more than 10x better than the second-best method HSCS. It is also interesting to see that \ours seriously overestimates the failure due to the suboptimal onion sampling. Such a bias is then sequentially reduced as more samples are collected.

\begin{figure}[]
  \centering
  %minipage
  % \begin{minipage}[t]{1\linewidth}
    % \centering
    % \vspace*{-0.1in}
    \includegraphics[width=1\linewidth]{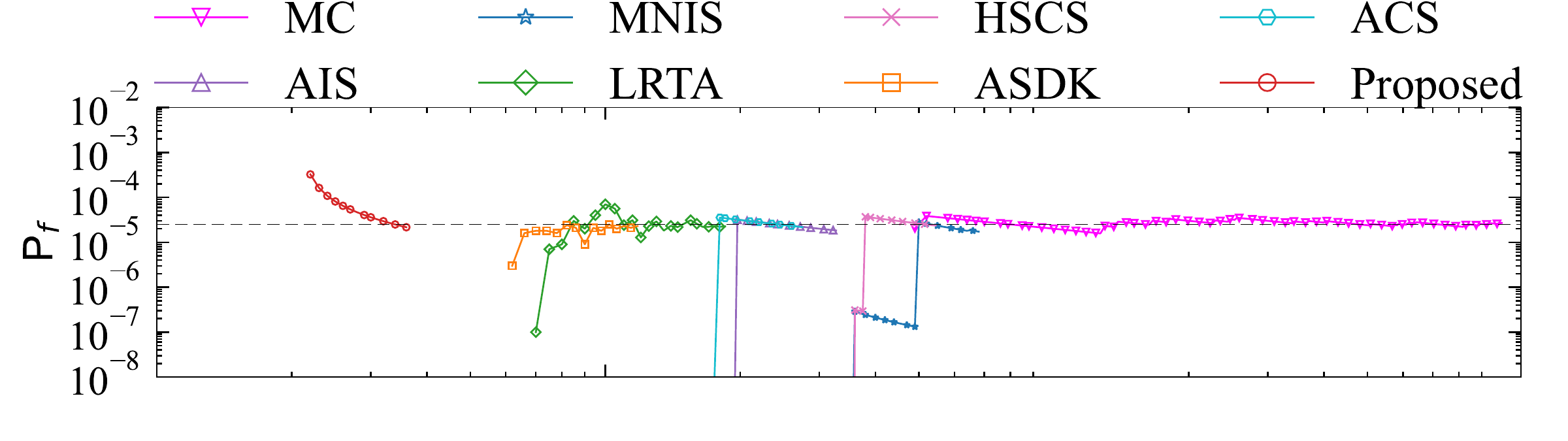}\\
    \vspace{-0.1in}
    \includegraphics[width=1\linewidth]{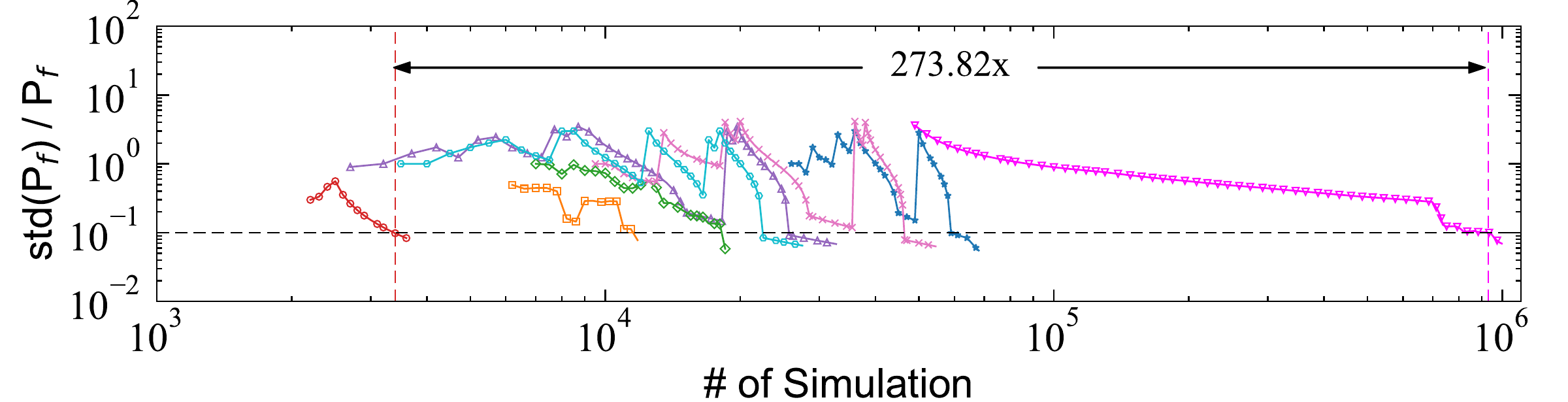}\\
    \vspace{-0.1in}
    \caption{$P_f$ and FOM on 569-dimensional SRAM column}
    \label{case4_exp}
  % \end{minipage}
  \vspace{-0.2in}
\end{figure}
\subsection{569-Dimensional SRAM Column Circuit}
To validate \ours in practical scenarios, we work on a commercial SRAM array solution with 528 transistors in the design to form bit-cell arrays, sense amplifiers, and power paths. Based on the transistor type (PMOS/NMOS), gate length, and gate width, each transistor will associate with 0-3 variational parameters (\ie mobility, oxide thickness, and saturation velocity) based on the BSIM model.
With a BSIM4 model, this leads to 569 variation parameters. 
% where each transistor instance is instantiated based on a model and instantiation parameters, such as the transistor type (PMOS/NMOS), the gate length, and the gate width, and each transistor has a random number between 0 and 3 associated with it. 
% The random numbers correspond to various parameters that affect the transistor behavior, such as mobility, oxide thickness, saturation velocity, etc.
% we further increase the number of SRAM bit cells in the SRAM column, which leads to an SRAM circuit with 569 variation parameters. 
The delay time of read/write of the SRAM acts as the output metric $\y$. 
% We run the same yield estimation algorithm on the 569-dimensional circuit.
The convergence dynamic is shown in \Figref{case4_exp} with results in \Tabref{AllCaseTable}.
Similarly, the results of \ours are significantly better than the competitors, showing a 273.82x speedup over MC, which is almost 3.5x faster than the second-best method ASDK. Most importantly, the estimation results are very close to the ground truth, with a relative error of 0.25\%. Again, the initial overestimation remains for \ours.

% The numerical results and detailed $P_f$ evolution are shown in  and . 
% The MC method achieves the ground truth of failure rate 2.50e-5 using 931000 simulations. 
% As shown in \Figref{case4_exp}, only after long importance sampling, are other IS-based methods able to discover the correct failure regions and their $P_f$ increases to the peak. In contrast, due to the powerful proposed pre-sampling approach, \ours discovers the correct failure regions and its $P_f$ achieves the peak right after the pre-sampling state, and it converges to the ground truth rapidly. 
% \ours achieves a fail rate of 2.49e-5 using 3400 simulations, exhibiting a relative error of 0.25\%, which outperforms others in terms of accuracy and efficiency. In comparison, MNIS, HSCS, AIS, ACS, LRTA and ASDK uses 59000, 46500, 25700, 22500, 18500 and ... simulations respectively to achieve a fail rate with the larger relative error. Compared with MNIS, \ours can achieve a up to 17.35x more accurate and 69.32x speedup. 

\begin{figure}[]
  \centering
  \vspace*{-0.1in}
  \includegraphics[width=1\linewidth]{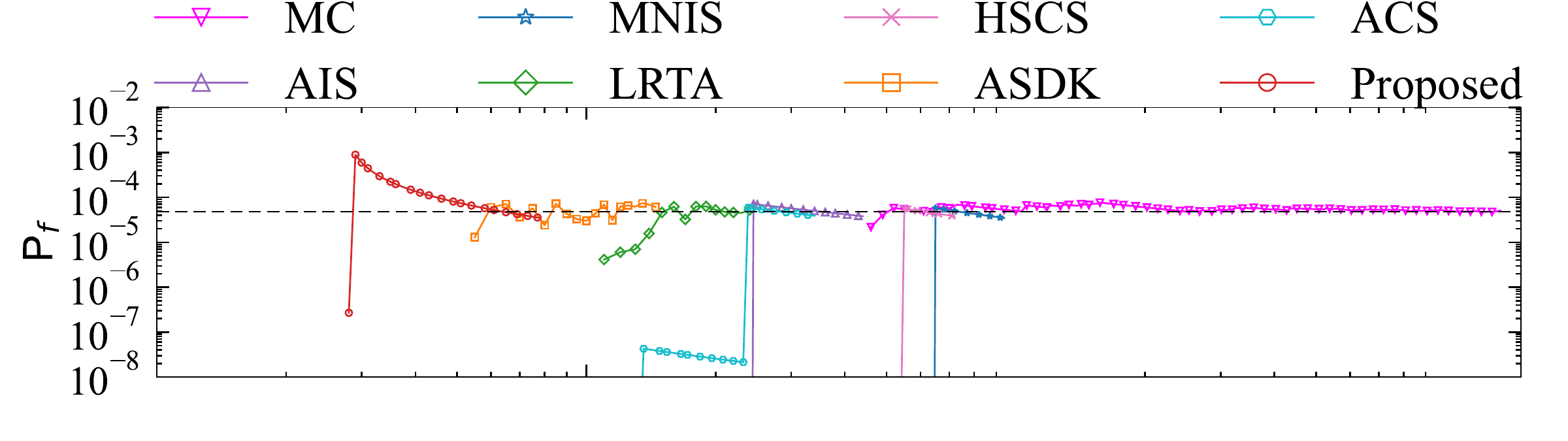}\\
  \vspace{-0.1in}
  \includegraphics[width=1\linewidth]{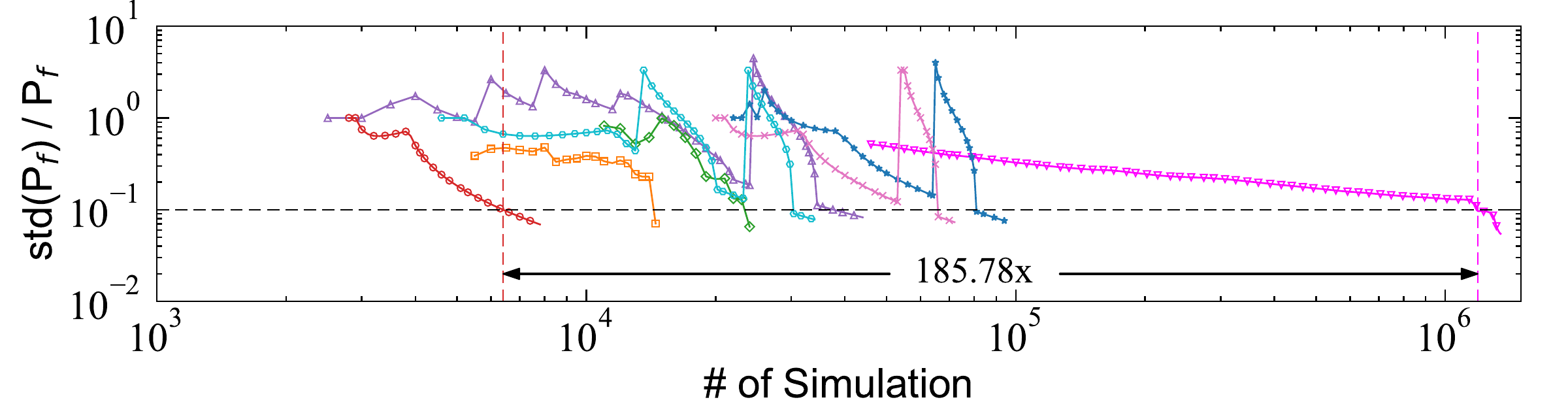}\\
  \vspace{-0.1in}
  \caption{$P_f$ and FOM on 1093-dimensional SRAM column}
  \label{case5_exp}
  \vspace{-0.1in}
\end{figure}
\subsection{1093-Dimensional SRAM Column Circuit}
We further increase the dimensions of the problem by using a detailed BSIM5 model, learning to 1093 variation parameters.
% of the circuit our problem by adding more bit cells, leading to an even more challenging SRAM column circuit with 1093 variation parameters.
As far as we are aware, no previous published work has ever attempted to estimate the yield of such a high-dimensional problem.
% and no former researcher implements yield estimation experiment on such a high-dimensional setting. 
Similarly, the delay time of read/write is used as the output metric. We run 1,189,000 simulations using the MC method and obtain the ground-truth failure rate, 4.80e-5. The results for the competing methods are shown in \Figref{case5_exp} and \Tabref{AllCaseTable}.
% The total experimental numerical results and detailed process are illustrated in \Tabref{AllCaseTable} and \Figref{case5_exp}.
The superior \ours is consistent, although this time, the improvement over the second-best method ASDK is 2.2x, not as significant as in the previous experiments. Nevertheless, the ASDK shows the largest error of 27.08\% among all methods. 

% It can be obviously concluded that \ours is still able to get the failure rate with highest accuracy and efficiency among other 6 methods, which uses merely 6400 runs of simulation and exhibits a estimated result with a 2.71\% relative error. Our method finds the correct failure region after one round of importance sampling, while other IS-based methods takes many rounds. The surrogate methods converge slower than \ours because they need a large number of training data.
% In comparison with MNIS, \ours can reach a up to 12.66x speedup and is 4.55x more accurate.

% \begin{table}
%   \centering
%       \caption{Ablation experiment of pre-sampling method}
%       \label{ablation}
%       \vspace*{-0.1in}
%       \begin{tabular}{c|c|c|c|c}
%          \midrule
%           & AIS&  AIS+  & ACS & ACS+ \\
%         \midrule
%           \# of IS & 13024 & 10854 & 12100 & 9600 \\
%           Rel. error & 7.98\% & 6.79\% & 6.59\% & 4.77\%  \\
%       \bottomrule
%     \end{tabular}
%     \vspace*{-0.15in}
%   \end{table}
\begin{table}
\centering
  \caption{Ablation experiment of pre-sampling method}
  \label{ablation}
  \vspace*{-0.1in}
  \begin{tabular}{c|ccc|ccc}
    \toprule
    & AIS & AIS+ & Impro. & ACS & ACS+ & Impro. \\
    \midrule
    Rel. error & 7.98\% & 6.79\% & 1.18x & 6.59\% & 4.77\% & 1.38x  \\
    \# of IS & 13024  & 10854  & 1.20x & 12100 & 9600 & 1.26x \\
    \bottomrule
  \end{tabular}
  \vspace*{-0.2in}
\end{table}

\subsection{Ablation Study}
% \begin{table}
%   \centering
%       \caption{Ablation experiment of pre-sampling method}
%       \label{ablation}
%       \vspace*{-0.1in}
%       \begin{tabular}{c|c|c|c|c|c|c}
%         \toprule
%            & \multicolumn{2}{|c|}{Origin method} & \multicolumn{2}{|c|}{With ours}  & \multicolumn{2}{|c}{Improvment}\\
%          \midrule
%           & Rel. error&  \# of IS  & Rel. error & \# of IS & Speedup & Acc.\\
%         \midrule
%                 AIS & 7.98\% & 13024 & 6.79\% & 10854 & & \\
%               ACS & 6.59\% & 12100 & 4.77\% & 9600 & & \\
%       \bottomrule
%     \end{tabular}
%     \vspace*{-0.15in}
%   \end{table}
% \subsection{Ablation Study}
% The onion sampling is the key contribution of this work.
We validate the usefulness of the onion sampling by equipping the classic IS methods, namely, AIS and ACS ,as their pre-sampling procedure and compare with their original versions.
% their results with their original LHS sampling.
The experiments are conducted on the same 108-dimensional SRAM experiment for its fast simulation speed.
1700 samples are given for all methods as their initial sampling budget, and the results are shown in \Tabref{ablation}, which shows about 20\% improvement in accuracy speedup with our onion sampling (AIS+ and ACS+).

% We give both method 1000 samples to start with and let them run for 1000 iterations. \help{confirm the number of iterations}.
% 
% We can clearly see that, using hyperspherical pre-sampling AIS and ACS run 11100 and 9300 simulations to achieve the fail rate of 4.74e-5 and 5.68e-5 respectively, and that, using our proposed pre-sampling method AIS and ACS run 10030 and 8000 to achieve the fail rate of 5.31e-5 and 5.45e-5 respectively. It can be concluded that the proposed pre-sampling method does help discover promising failure regions and accelerate the speed of convergence of fail probability.
% 

% To verify the proposed pre-sampling method's contribution to the performance of \ours, we conduct the ablation experiment on the 
% ACS and AIS use the same pre-sampling algorithm, named hyperspherical pre-sampling, which draws samples on hyperspheres with increasing radius in the variation parameter space. We replace AIS and ACS's pre-sampling method with our proposed method, and implements the modified AIS and ACS on the circuit. The circuit setup is as the same as that of the Section \ref{108dim_exp}.
% To evaluate the method fairly, the modified pre-sampling procedure are fixed to run as the same number of simulation as that of the origin pre-sampling procedure.
% The real failure probability of the circuit is set as 5.01e-5. 

\subsection{{Robustness Study}}
To further assess the robustness of our method, we conduct a robustness study on the 108-dimensional SRAM circuit by running experiments 10 times with random initializations. The final estimations with relative errors larger than 50\% are marked fail. The statistical results for successful runs are shown in \Tabref{robustness}. As we expect, the surrogate methods, LRTA and ASDK, suffer from great instability with more than 5 times fail, whereas the IS methods are more stable. In contrast, \ours shows the best performance for all metrics.

% \help{what is the final estimation?}
% \vspace{-0.2in}
\begin{table}
  \vspace{-0.15in}
  \centering
  \caption{Robustness test on 108-dimensional SRAM column}
  \label{robustness}
  \vspace{-0.1in}
    \begin{adjustbox}{width=1\columnwidth,center}
      \begin{tabular}{c|cccccccc}
        \toprule
         Method & MNIS & HSCS & AIS & ACS & LRTA & ASDK & Proposed \\
        \midrule
         Avg. RE  & 16.00\% & 8.90\% & 6.64\% & 6.45\%  & 12.85\% & 10.18\% & \bf{1.91\%} \\
        % & Mean \# of sim.      & 699000.00 & 50883.33 & 31340.00 & 15714.29 & 14071.43   & 13440.00  &  & 5300.00 \\
         Avg. speedup & 13.73x & 22.30x & 44.48x & 49.68x & 52.01x & 75.98x & \bf{131.89x} \\
       \midrule
        \# Fail & 4/10 & 4/10 & 3/10 & 3/10 & 5/10 & 9/10 & \bf{1/10} \\
      \bottomrule
    \end{tabular}
\end{adjustbox}
\vspace*{-0.2in}
\end{table}

% \begin{table*}
%   \centering
%       \caption{Numerical results on 108-dimensional SRAM column}
%       \begin{tabular}{c|c|cccccccc}
%         \toprule
%          & Results & MC & MNIS & HSCS & AIS & ACS & LRTA & ASDK & Proposed \\
%         \midrule
        
%         \multirow{2}{*}{Successful runs} 
%         % & Mean failure prob.   & 5.01e-5 & 4.21e-5 & 4.56e-05 & 4.67e-5 & 5.33e-5  & 4.37e-05 &  & 5.10e-5 \\
%         & Mean relative error  & - & 16.00\% & 8.90\% & 6.64\% & 6.45\%  & 12.85\% & 10.18\% & 1.91\% \\
%         % & Mean \# of sim.      & 699000.00 & 50883.33 & 31340.00 & 15714.29 & 14071.43   & 13440.00  &  & 5300.00 \\
%         & Mean speedup & 1x & 13.73x & 22.30x & 44.48x & 49.68x & 52.01x & 75.98x & 131.89x \\
%        \midrule
      
%        \multirow{1}{*}{Failed runs} 
%         & Failed times & N/A & 4/10 & 4/10 & 3/10 & 3/10 & 5/10 & 9/10 & 1/10 \\
%       \bottomrule
%     \end{tabular}
%   \end{table*}

\section{Conclusion}
% \Wei{In this paper, a novel yield estimation framework is proposed for high-dimensional SRAM circuits.} 
We generalize NM to optimal manifold and propose \ours to deliver the SOTA yield estimation. Limitation of \ours includes the implicit assumption of one failure boundary. Also, the NSF might need tuning for different problems, even though changing the NN structure in NSF does not have a significant influence on all the conducted experiments.

% At the same time, too little training data will lead to a large deviation between the  distribution gained by NSF and the real distribution.

% \section*{Acknowledgment}

\vspace{-0.1in}
\bibliographystyle{IEEEtran}
\bibliography{BYA}

\end{document}